%% file: main.tex
\documentclass[letterpaper]{article} % DO NOT CHANGE THIS
\usepackage{aaai2027}  % DO NOT CHANGE THIS
\nocopyright
\usepackage[hyphens]{url}  % DO NOT CHANGE THIS
\usepackage{graphicx} % DO NOT CHANGE THIS
\usepackage{natbib}  % DO NOT CHANGE THIS AND DO NOT ADD OPTIONS
\usepackage{caption} % DO NOT CHANGE THIS AND DO NOT ADD OPTIONS
\usepackage{amsmath,amssymb}
\usepackage{booktabs}
\usepackage{multirow}
\usepackage{tabularx}
\usepackage{float}
\newcolumntype{Y}{>{\raggedright\arraybackslash}X}
\usepackage{tikz}
\usetikzlibrary{patterns}
\definecolor{plotblue}{HTML}{0077BB}
\definecolor{plotorange}{HTML}{EE7733}
\tikzset{
  fastbar/.style={draw=black, line width=0.25pt, fill=plotblue!55,
    postaction={pattern=north east lines, pattern color=black!45}},
  oursbar/.style={draw=black, line width=0.25pt, fill=plotorange!85}
}
\title{Vid2WAM: Distilling Video Diffusion Priors into World Action Models}

\author{
Chenhao Qiu\textsuperscript{\rm 1}\equalcontrib,
Ruixiang Wang\textsuperscript{\rm 2}\equalcontrib\thanks{Project lead.},
Runyi Zhao\textsuperscript{\rm 2}\equalcontrib,
Sixu Lin\textsuperscript{\rm 2},
Songen Gu\textsuperscript{\rm 1},\\
Shufeng Nan\textsuperscript{\rm 1},
Guiliang Liu\textsuperscript{\rm 2},
Kui Jia\textsuperscript{\rm 2},
Yanwei Fu\textsuperscript{\rm 1,\rm 3},
Simo Wu\textsuperscript{\rm 1}\corresponding
}
\affiliations{
\textsuperscript{\rm 1}Fudan University\\
\textsuperscript{\rm 2}The Chinese University of Hong Kong, Shenzhen\\
\textsuperscript{\rm 3}Shanghai Innovation Institute
}

\begin{document}

\maketitle

%%%%%%%%%%%%%%%%%%%%%%%%%%%%%%% text

\input{sections/0_abs}
\input{sections/1_intro}

\input{sections/2_rel}
\input{sections/3_method}

\input{sections/4_exp}

\input{sections/5_conclusion}

%%%%%%%%%%%%%%%%%%%%%%%%%%%%%%% text

\bibliography{main}

\clearpage
\appendix

\begingroup
\setcounter{secnumdepth}{1}
\setcounter{table}{0}
\setcounter{figure}{0}
\setcounter{equation}{0}
\renewcommand{\thetable}{S\arabic{table}}
\renewcommand{\thefigure}{S\arabic{figure}}
\renewcommand{\theequation}{S\arabic{equation}}

\makeatletter
\setlength{\@fptop}{0pt}
\setlength{\@fpsep}{10pt}
\setlength{\@fpbot}{0pt plus 1fil}
\setlength{\@dblfptop}{0pt}
\setlength{\@dblfpsep}{12pt}
\setlength{\@dblfpbot}{0pt plus 1fil}
\makeatother
\setcounter{topnumber}{4}
\setcounter{bottomnumber}{2}
\setcounter{totalnumber}{6}
\setcounter{dbltopnumber}{3}
\renewcommand{\topfraction}{0.95}
\renewcommand{\bottomfraction}{0.80}
\renewcommand{\floatpagefraction}{0.75}
\renewcommand{\dbltopfraction}{0.95}
\renewcommand{\dblfloatpagefraction}{0.75}
\renewcommand{\textfraction}{0.05}

\input{supp/sections/teacher_idm}
\input{supp/sections/implementation_detail}
\input{supp/sections/task_selection_sensitivity}
\input{supp/sections/realworld_evaluation}
\input{supp/sections/robotwin_analysis}

\clearpage
\endgroup

\end{document}

%% file: sections/0_abs.tex
\begin{abstract}
% World Action Models (WAMs) improve robot policy learning by jointly modeling future visual dynamics and actions. However, their scalability and generalization remain constrained by their reliance on costly expert demonstrations. We therefore ask whether future supervision for WAMs must come from real robot trajectories. In this paper, we propose \textbf{Vid2WAM}, an offline distillation framework that transfers visual dynamics priors from a large video foundation model into a compact WAM student. Given an observation and language instruction, the teacher generates task-conditioned future rollouts as visual targets, while an inverse dynamics model infers the corresponding pseudo-actions as action targets. To reduce interference from pseudo-action noise during policy learning, we introduce source-aware residual action adaptation, which routes synthetic supervision through a residual adapter while using real trajectories to anchor the shared action head. At inference time, both the video teacher and inverse dynamics model are discarded, leaving only the compact WAM student for deployment. Experiments in simulation and real-world demonstrate that Vid2WAM improves both generalization to novel tasks and efficiency under limited expert demonstrations while preserving low-latency inference.

World Action Models (WAMs) improve robot policy learning by jointly modeling future visual dynamics and actions. However, their scalability and generalization remain constrained by their reliance on costly expert demonstrations. We challenge this by asking whether future supervision for WAMs must originate from target-task expert trajectories. In this paper, we propose \textbf{Vid2WAM}, an offline distillation framework that transfers visual diffusion priors from a large video foundation model into a compact WAM student. Given an observation and language instruction, Vid2WAM distills supervision through two complementary channels: task-conditioned future rollouts directly supervise the student's future prediction branch, while an inverse dynamics model recovers embodiment-specific pseudo-actions for action learning. To robustly integrate synthetic and real supervision, we introduce  \textit{source-aware residual action adaptation} that learns source-specific corrections around a shared action backbone and mitigates interference from noisy pseudo-actions. During inference, both the video teacher and inverse dynamics model are discarded, leaving only the WAM student for efficient deployment. Simulation and real-world experiments demonstrate that Vid2WAM improves novel-task generalization and data efficiency under limited expert demonstrations while preserving low-latency inference.

\end{abstract}

%% file: sections/1_intro.tex
\section{Introduction}

\begin{figure}[!t]
    \centering
    \includegraphics[width=1\linewidth]{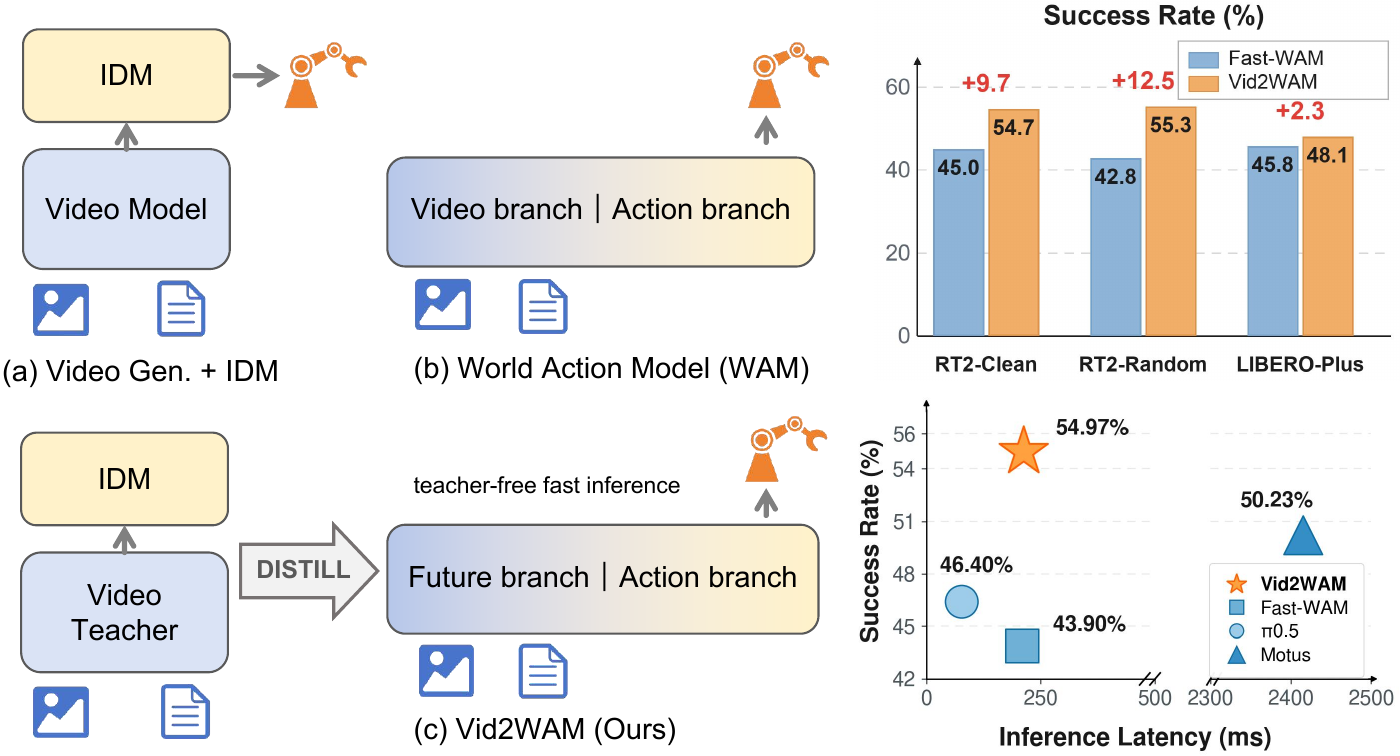}
\caption{\textbf{Left:} Comparison of video-based robot policy paradigms. (a) Video generation plus IDM performs costly online rollout and action recovery. (b) WAM jointly learns future prediction and action generation from real video--action trajectories. (c) Vid2WAM distills teacher-generated futures and IDM pseudo-actions into a compact WAM for teacher-free inference. \textbf{Top right:} Vid2WAM consistently improves novel-task success rates over Fast-WAM. \textbf{Bottom right:} Vid2WAM achieves the highest novel-task success rate while retaining low inference latency.\label{fig:placeholder} }
\vspace{-0.15in}

\end{figure}

% Building robot policies that generalize across tasks, objects, and environments is a central goal of robot learning. One prominent route is vision--language--action (VLA) modeling, which adapts Internet-pretrained vision--language models for action prediction and thereby transfers broad semantic and linguistic knowledge into control~\cite{zitkovich2023rt2,kim2025openvla,black2024pi0}. A complementary route builds robot policies around visual foresight. Video-based policies and World Action Models (WAMs) incorporate future prediction into action learning, providing spatiotemporal supervision over motion, interaction, and task progression beyond static semantic supervision and instantaneous action labels~\cite{du2023unipi,wu2024gr1,hu2025vpp,xu2025imaginative,kim2026cosmospolicy,ye2026dreamzero}.

Building robot policies that generalize across tasks, objects, and environments remains a central goal of embodied intelligence. Recent progress has followed two complementary directions. Vision--language--action (VLA) models leverage Internet-scale vision--language pretraining to transfer broad semantic and linguistic knowledge into control~\cite{zitkovich2023rt2,kim2025openvla,black2024pi0}. Meanwhile, video-based policies and World Action Models (WAMs) incorporate future prediction into policy learning, providing spatiotemporal supervision over motion, interaction, and task progression beyond static semantic supervision and instantaneous action labels ~\cite{du2023unipi,wu2024gr1,hu2025vpp,xu2025imaginative,kim2026cosmospolicy,ye2026dreamzero}.

% By coupling future modeling with action learning, WAMs expose the policy to temporally structured supervision from future observations. Recent studies show 
% that such predictive objectives can shape 
% dynamics-aware representations and improve action 
% learning even when explicit future rollouts are 
% not required at 
% deployment~\cite{li2025uva,xu2025imaginative,yuan2026fastwam}. 
% However, most WAMs still obtain task-specific future and action targets from 
% paired expert video-action trajectories, tying the supervision to the coverage of costly demonstrations. This raises our central 
% question: can the predictive priors of a large video-based model 
% provide future supervision beyond recorded target-task trajectories and 
% be distilled into an efficient WAM policy?

A key advantage of WAMs stems from their future-predictive supervision. Rather than learning only from action labels, they are optimized to anticipate future visual dynamics, providing dense temporal learning signals that improve action representations even when explicit future rollouts are not required during deployment~\cite{li2025uva,xu2025imaginative,yuan2026fastwam}. However, most existing WAMs implicitly share one assumption: \textit{future supervision must be observed in task-specific expert demonstrations}. Both future visual targets and action labels are obtained from paired robot trajectories collected through simulated data collection or teleoperation. 
Thus, the predictive capability of WAMs remains closely tied to  scale, diversity, and cost of robot demonstrations, making it difficult to generalize beyond distributions covered by recorded target-task trajectories.

We thus argue that the assumption above is unnecessarily restrictive. Recent video foundation models have acquired remarkably rich knowledge of object motion, human-object interactions, and long-horizon visual dynamics. After adaptation with limited embodiment-specific data, they can be queried with diverse initial observations and language instructions to synthesize task-conditioned rollouts for new tasks and environments~\cite{jang2025dreamgen,wang2026eva}.
Unlike expert demonstration collection, this process does not require a complete target-task trajectory, allowing the conditioning distribution to be expanded using simulator snapshots, inexpensive real-world observations, or controllable image synthesis. An inverse dynamics model (IDM) can further recover embodiment-specific pseudo-actions from the generated rollouts~\cite{jang2025dreamgen,ye2025lapa}. However, treating these rollouts only as pseudo-labeled action data creates an information bottleneck: the teacher's predictive knowledge is compressed through an imperfect action-recovery model, while the generated futures themselves are discarded as direct supervision. Moreover, ambiguous transitions, generation artifacts, and physically inconsistent motion can induce systematic errors in the recovered actions~\cite{wang2026eva}. These limitations motivate transferring the generated rollouts through complementary visual and action channels.

This paper presents \textbf{Vid2WAM}, a framework for distilling a large video foundation model into an efficient WAM student. Given an initial observation and language instruction, an embodiment-adapted teacher generates a future rollout. We decode this rollout into pixels and re-encode it into the student's latent space, providing representation-compatible supervision for the future prediction branch. In parallel, an IDM recovers embodiment-specific pseudo-actions that supervise action learning. To mitigate the interference from pseudo-action noise, we further introduce \textit{source-aware residual action adapters}, which provide source-specific residual corrections around a shared action backbone. At inference time, the video teacher and IDM are removed, leaving only the compact WAM student for direct action generation.

% \textit{Our Vid2WAM reframes the role of video foundation models in robot learning}. Rather than treating them as generators of additional pseudo trajectories, we view them as scalable providers of future supervision that can replace expensive robot rollouts during training while preserving efficient student-only deployment. This perspective decouples world-model learning from the availability of robot demonstrations and substantially expands the supervision available for learning predictive robotic policies.

Our contributions are threefold. First, we rethink the source of future supervision for World Action Models and formulate a video-foundation-model-to-WAM distillation framework, using generated futures from video foundation models as direct supervision beyond recorded target-task expert trajectories. Second, we introduce a dual-channel, source-aware transfer strategy that jointly distills future visual dynamics and IDM-inferred actions, with residual adapters reducing interference from noisy pseudo-actions. Third, we demonstrate in simulation and real-world experiments that Vid2WAM improves expert-demonstration efficiency and novel-task generalization, while retaining efficient student-only inference.

% Our contributions are threefold. First, we formulate 
% video-foundation-model-to-WAM distillation, using generated futures as direct supervision rather than only as pseudo-action data. Second, we introduce source-aware transfer that jointly distills future visual dynamics and IDM-inferred actions, with residual adapters reducing interference from noisy pseudo-labels. Third, we demonstrate that Vid2WAM improves data efficiency and novel-task generalization in simulation and the real world, while retaining efficient student-only inference.

%% file: sections/2_rel.tex
\section{Related Work}

\noindent \textbf{World Models for Robot Control.}
Video models provide broad priors over motion, interaction, and temporal task structure. One line of work uses these priors explicitly at deployment, generating visual subgoals or future rollouts and converting them into actions through inverse dynamics, goal-conditioned control, motion tracking, or test-time action selection
~\cite{du2023unipi,du2024vlp,zhou2024robodreamer,liang2025dreamitate,collins2025amplify,zhou2026tau0wm}.
Although effective for generalization, this paradigm incurs generation latency and exposes control to rollout errors. A related line of work jointly models future observations and robot actions, using future prediction to provide temporally structured supervision for policy learning
~\cite{wu2024gr1,cheang2024gr2,hu2025vpp,li2025uva,zhu2025uwm}.
Fast-WAM shows that video co-training can remain beneficial even when future generation is removed at inference~\cite{yuan2026fastwam}, while Efficient-WAM reduces  cost of retaining future imagination~\cite{li2026efficientwam}. 
These methods nevertheless derive both future and action supervision primarily from recorded robot trajectories. Vid2WAM instead decouples future supervision from target-task expert demonstrations by leveraging a video foundation model as an offline generator of task-conditioned futures, which are distilled into a compact WAM without requiring video generation at deployment.

% These methods nevertheless derive future and action targets primarily from recorded robot trajectories. Vid2WAM instead uses a large video model as an offline source of future supervision beyond the available expert trajectories.

% \noindent \textbf{Generative Models for Synthetic Robot Data.} 

\noindent \textbf{Video-Generated Supervision for Robot Learning}. 
Early generative augmentation methods edit objects, backgrounds, and other visual factors in recorded demonstrations while reusing their actions
~\cite{chen2023genaug,yu2023rosie}. More recent work also synthesizes or edits initial observations, expanding the states from which robot experience can be generated~\cite{kim2026robocurate}. DreamGen provides the closest data-centric pipeline: it adapts a video model to a target embodiment, generates task-conditioned rollouts, and recovers pseudo-actions using an inverse dynamics or latent action model
~\cite{jang2025dreamgen,ye2025lapa}. 
%
% In this formulation, generated frames and recovered actions form synthetic demonstrations, but the generated futures do not directly supervise a student's future-prediction objective. Consequently, action learning still depends on labels recovered through an imperfect action model. 
%
In these pipelines, generated frames and recovered actions form synthetic demonstrations, but the generated futures do not directly supervise a student's future-prediction objective. This action-only transfer creates an information bottleneck, compressing rich predictive knowledge into imperfect pseudo-labels before it reaches the student policy.
Prior work mitigates this issue by improving the executability of generated videos~\cite{wang2026eva} or by avoiding synthetic action supervision and routing geometric targets to the visual backbone~\cite{chen2026gra}. 
Vid2WAM instead treats each generated rollout as dual supervision: future dynamics directly supervise world modeling, while IDM-derived actions provide embodiment-specific action grounding through source-aware residual adaptation.

% Vid2WAM instead uses each generated rollout in two complementary ways: future latents directly supervise world modeling, while IDM-derived actions provide embodiment-specific action grounding through source-aware residual adaptation.

\noindent \textbf{Distilling Foundation-Model Priors into Robot Policies.}
Knowledge distillation transfers capabilities from a large teacher into a smaller student~\cite{hinton2015distilling}, and subsequent work has extended this principle from output imitation to teacher-generated data and intermediate supervision
~\cite{hsieh2023distilling,mukherjee2023orca,li2023dreamteacher}. In robotics, prior methods distill planner-generated experience or foundation-model representations into deployable policies
~\cite{ha2023scaling,shang2024theia}. More closely related, S-VAM self-distills multi-step generated videos into one-step geometric and semantic foresight, while PFD distills action corrections induced by privileged recorded futures
~\cite{yan2026svam,fang2026pfd}. CKT-WAM transfers intermediate context between heterogeneous WAMs, whereas Efficient-WAM compresses the computation required for future imagination
~\cite{jiang2026cktwam,li2026efficientwam}. 
Vid2WAM addresses a different transfer setting. Rather than distilling predictions between robot policies or WAMs, it treats a video foundation model as an external source of future supervision. Each generated rollout provides complementary supervision for both future prediction and action learning, after which the video teacher and inverse dynamics model are discarded, leaving only the compact WAM student for deployment.

%
% Vid2WAM addresses a different transfer setting: an external video foundation model is queried offline beyond recorded target-task trajectories, and each generated rollout supervises both future prediction and action learning. The video teacher and IDM are removed after training, leaving only the compact student policy at deployment.

%% file: sections/3_method.tex
\section{Method}
\label{sec:method}

\begin{figure*}[t]
    \centering
    \includegraphics[width=0.95\textwidth]{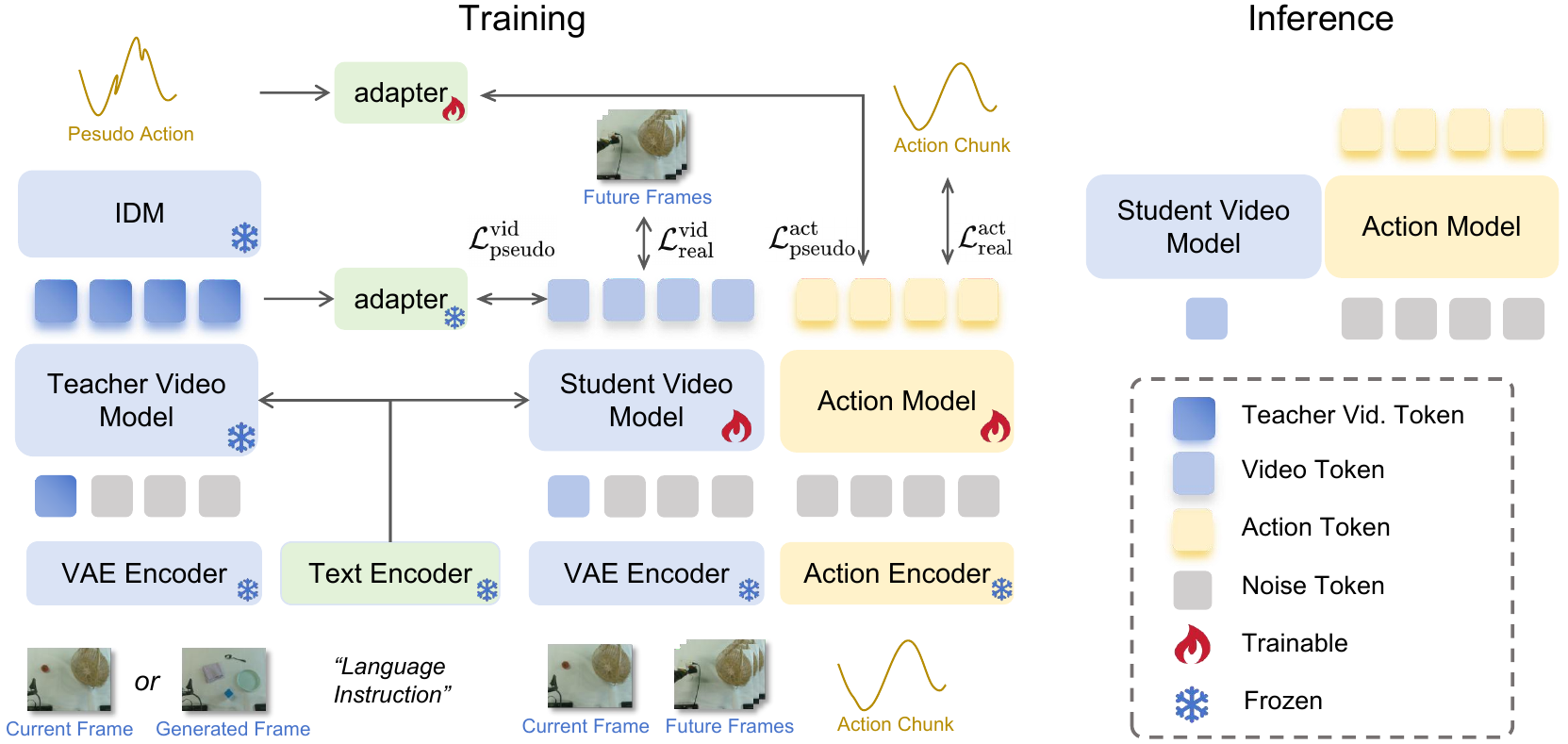}
    \caption{\textbf{Overview of Vid2WAM.} Vid2WAM distills future prediction priors from a pretrained video diffusion model into a compact World Action Model (WAM). During training, the teacher generates future videos and IDM-derived pseudo actions supervise the student WAM, while only the student is used for deployment. At inference, a single current-frame pass through the shared video backbone produces latent world features for direct action prediction.}
    \label{fig:method}
\end{figure*}

\noindent \paragraph{Problem Formulation and Overview}
% \label{sec:problem_setup}
Given a current visual observation $o_0$, proprioceptive state $e_0$,
and language instruction $\ell$, we aim to learn a policy that predicts
an executable action chunk $a_{1:H}$. 
% We denote the policy condition as
% $c=(o_0,e_0,\ell)$. 
Following the World Action Model (WAM) paradigm,
the student is jointly trained on an action predictor and an auxiliary
future-video predictor through a shared video-action representation.

As illustrated in Fig.~\ref{fig:method}, Vid2WAM transfers the predictive prior of a large video teacher into the student using both real demonstrations and offline teacher-generated rollouts. Real demonstrations provide expert actions and observed future-video supervision, whereas teacher rollouts provide generated future-video latents and IDM-derived pseudo-actions. The video teacher and IDM are used only offline; at deployment, the teacher, IDM, auxiliary future-video prediction head, and pseudo-source adapters are discarded, leaving the student WAM to efficiently generate action chunks from the current policy condition.

\subsection{Teacher-Derived Distillation Targets}
\label{sec:offline_teacher}

We convert the video teacher's generative prior into two complementary
forms of student-compatible supervision: future-video latent targets and
IDM-derived pseudo-action targets.

\paragraph{Video teacher adaptation.}
We instantiate the video teacher $T_{\psi}$ with a large pretrained video
diffusion model and fine-tune it using only a limited
set of demonstration videos. This lightweight adaptation
specializes the teacher to the robot's appearance and embodiment-specific
physical constraints, while preserving the motion-prediction and instruction-following capabilities inherited from large-scale pretraining. Consequently,
a small amount of in-domain demonstration data is sufficient to elicit
robot-aware rollout generation. Given an initial observation $o_0$ and
language instruction $\ell$, the adapted teacher generates
\begin{equation}
    \hat{v}_{1:T}=T_{\psi}(o_0,\ell).
\end{equation}
After adaptation, the teacher is frozen and used exclusively for offline
rollout generation.

\paragraph{Inverse dynamics model training.}
We train an inverse dynamics model (IDM) to recover robot motion from
visual trajectories:
\begin{equation}
(\hat{a}_{1:H}, \hat{e}_{1:H})=\mathrm{IDM}(\hat{v}_{1:T}),
\end{equation}
where $\hat{a}_{1:H}$ and $\hat{e}_{1:H}$ denote the predicted action chunk
and robot state sequence respectively. The IDM is supervised using
action-labeled robot trajectories, but does not require language instructions or
task-specific semantic annotations. It can therefore be trained using
task-agnostic interaction data collected with the same robot embodiment and
compatible observation setup, without requiring expert demonstrations for the
target tasks. Once trained, the IDM is frozen and used only as a
labeler for teacher-generated rollouts.

\paragraph{Pseudo-data generation and labeling.}
Vid2WAM keeps the real demonstration set fixed and collects an additional
pool of initial states.
The frozen video teacher produces visual rollouts from these initial states,
which are re-encoded by the student VAE into student-space
future latents $\hat{z}_{1:T}$. The  IDM then recovers the corresponding
pseudo-action chunks $\hat{a}_{1:H}$ and robot state sequences
$\hat{e}_{1:H}$ from the generated rollouts. The resulting tuples
$(o_0,\ell,\hat{z}_{1:T},\hat{a}_{1:H},\hat{e}_{1:H})$
are stored in an offline distillation buffer. The future latents supervise
the student video branch, while the IDM-derived labels ground the
generated motion in the robot action space.

\subsection{Source-Aware Student Training}
Vid2WAM jointly trains the student on real demonstrations and offline
teacher-generated pseudo rollouts. Both domains share the video and action
backbones. 
\paragraph{Source-specific residual adaptation.}
\label{sec:adapter}
Directly mixing expert actions with IDM-recovered pseudo-actions can introduce
interference because the latter contain errors from video generation and inverse
dynamics. We therefore place lightweight residual adapters before and after the
shared action backbone, with separate parameters for the real and pseudo
domains. For an action token $x$ from domain
$d\in\{\mathrm{real},\mathrm{pseudo}\}$, an adapter computes
\begin{equation}
\label{eq:domain_adapter}
\operatorname{Ada}_{d}(x)=x+\alpha W_{\mathrm{up}}^{d}\sigma\!\left(W_{\mathrm{down}}^{d}\operatorname{LN}(x)\right),
\end{equation}
where $W_{\mathrm{down}}^{d}$ and $W_{\mathrm{up}}^{d}$ form a bottleneck
projection, $\sigma$ is a nonlinear activation, and $\alpha$ controls the
residual contribution. We initialize $W_{\mathrm{up}}^{d}$ to zero, so each
adapter initially implements an identity mapping. The shared representation
therefore remains the default path, while the residual branches provide
source-dependent corrections. Only the
real-domain adapter is retained at inference time.

\paragraph{Training objective.}
\label{sec:training_objective}
We use the flow-matching objective for action chunks and future-video
latents~\cite{lipman2022flow}. Let $y$ denote either
target, we sample Gaussian noise $\epsilon\sim\mathcal{N}(0,I)$ and $t\in(0,1)$, and construct
\begin{equation}
\label{eq:fm_noising}
y_t=(1-t)y+t\epsilon.
\end{equation}
For each domain $d$, the model predicts the
corresponding velocity field with flow-matching loss
\begin{equation}
\label{eq:fm_loss}
\mathcal{L}_{\mathrm{FM}}(y,d)=\mathbb{E}_{y,\epsilon,t}\!\left[\left\|f_{\theta}(y_t,t,c;d)-(\epsilon-y)\right\|_2^2\right].
\end{equation}
We group the four training terms into real- and pseudo-domain objectives:
\begin{equation}
\begin{aligned}
\mathcal{L}_{\mathrm{real}}
&=
\mathcal{L}_{\mathrm{FM}}(a,\mathrm{real})
+
\beta\mathcal{L}_{\mathrm{FM}}(z,\mathrm{real}),\\
\mathcal{L}_{\mathrm{pseudo}}
&=
\mathcal{L}_{\mathrm{FM}}(\hat{a},\mathrm{pseudo})
+
\beta\mathcal{L}_{\mathrm{FM}}(\hat{z},\mathrm{pseudo}),
\end{aligned}
\label{eq:domain_losses}
\end{equation}
where $(a,z)$ are the action and future-video targets from real
demonstrations, while $(\hat{a},\hat{z})$ are the corresponding targets from
teacher-generated distillation buffers. The complete training objective is
\begin{equation}
\mathcal{L}
=
\lambda_{\mathrm{real}}\mathcal{L}_{\mathrm{real}}
+
\lambda_{\mathrm{pseudo}}\mathcal{L}_{\mathrm{pseudo}},
\label{eq:total_loss}
\end{equation}
where $\beta$ balances action and video supervision, and
$\lambda_{\mathrm{real}}$ and $\lambda_{\mathrm{pseudo}}$ balance the two
training domains.

% \subsection{3.4 Action-Only Deployment}
% At inference time, Vid2WAM discards the teacher, IDM, distillation buffer,
% pseudo stream, pseudo-domain adapter, video DiT, and video head. The deployed
% policy uses only the action DiT and real-domain action adapter to predict action
% chunks, matching the low-latency execution pattern of the demonstration-only
% student while retaining the shared representation learned from offline future
% supervision.

%% file: sections/4_exp.tex
\section{Experiments}

% We organize the experiments around the questions each evaluation answers:
% \begin{itemize}
%     \item Can Vid2WAM improve a Fast-WAM-style student under a fixed
%     scarce-demonstration budget in the low-data regime? (Section~4.2)
%     \item Does video-teacher distillation improve robustness under perturbations
%     on LIBERO-Plus? (Section~4.3)
%     \item Does the same training signal support novel-task adaptation without
%     real action trajectories? (Section~4.4)
%     \item How do action-branch sharing and source-specific adapters affect
%     performance? (Section~4.5)
%     \item Does Vid2WAM preserve the low inference latency of its Fast-WAM
%     student after offline distillation? (Section~4.6)
% \end{itemize}

We investigate four questions: (1) how well existing pretrained
VLA and WAM policies transfer to novel tasks; (2) whether Vid2WAM improves performance on such tasks
without target-task expert trajectories; (3) whether it improves task success
and perturbation robustness under a fixed, limited budget of real
demonstrations; and (4) how its supervision signals and source-aware components
contribute to these gains. We conduct experiments on RoboTwin~2.0~\cite{chen2025robotwin2}, LIBERO~\cite{liu2023libero}, LIBERO-Plus~\cite{fei25libero-plus}, and a real-world bimanual platform.

\subsection{Experimental Setup}

\paragraph{Simulation Benchmarks.}
% We evaluate Vid2WAM on three complementary simulation benchmarks: LIBERO, LIBERO-Plus and RoboTwin 2.0. 
For RoboTwin 2.0, we evaluate bimanual manipulation for 50 tasks under both clean and randomized conditions, with 100 trials per task for every policy. For LIBERO, we use the Spatial, Object, Goal and
Long suites, each with 10 tasks and evaluate for 50 trials per task. LIBERO-Plus stresses robustness by applying
perturbations to LIBERO tasks along seven factors: camera, robot initial state,
language, lighting, background, sensor noise, and layout. We use the
protocol with trials on 500 variants per suite, uniformly covering all perturbations.

% \begin{table*}[!t]
% \centering
% \small
% \setlength{\tabcolsep}{4pt}
% \begin{tabular}{lccccc|ccccc}
% \toprule
% \multirow{2}{*}{Method}
% & \multicolumn{5}{c|}{Low-data}
% & \multicolumn{5}{c}{Novel} \\
% \cmidrule(lr){2-6} \cmidrule(lr){7-11}
% & Spatial & Object & Goal & Long & \textbf{Avg.}
% & Spatial & Object & Goal & Long & \textbf{Avg.} \\
% \midrule
% Motus
% & \textbf{93.0} & 94.4 & 85.0 & 78.4 & 87.7
% & 77.4 & 72.2 & 68.8 & 69.8 & 72.05 \\
% $\pi_{0.5}$
% & 91.2 & 95.6 & 84.4 & 76.4 & 86.9
% & 77.2 & \textbf{80.2} & 78.4 & \textbf{75.8} & 77.9 \\
% Fast-WAM
% & 89.6 & 97.8 & 82.8 & 78.8 & 87.25
% & 76.8 & 79.0 & 77.8 & 73.0 & 76.65 \\
% Ours
% & \textbf{93.0} & \textbf{98.6} & \textbf{87.6} & \textbf{79.4} & \textbf{89.65}
% & \textbf{79.4} & 79.4 & \textbf{79.2} & 75.2 & \textbf{78.3} \\
% \bottomrule
% \end{tabular}
% \caption{Results on LIBERO in the low-data and novel regimes.}
% \label{tab:libero}
% \label{tab:libero1}
% \end{table*}

\begin{figure*}[t]
\centering
\includegraphics[width=1.0\textwidth]{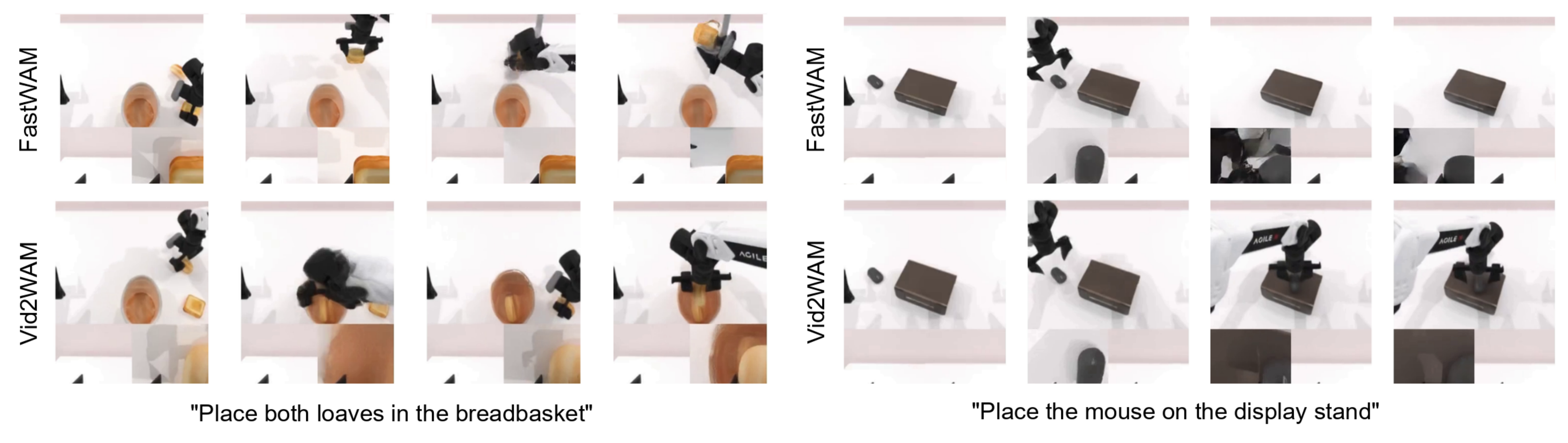}
\caption{Qualitative future-video predictions for novel tasks. For each task, Fast-WAM
(top) and Vid2WAM (bottom) receive the same initial observation and language
instruction; frames progress from left to right. Fast-WAM often fails
to predict coherent task progression, whereas Vid2WAM predicts future motions
that more consistently reach the instructed outcome.}
\label{fig:qualitative_video_predictions}
\end{figure*}

\begin{figure*}[h]
\centering
\begin{tikzpicture}[x=1cm,y=2.45cm]
\begin{scope}[xshift=0.4cm]
  \foreach \y/\lab in {0/0,0.2/.2,0.4/.4,0.6/.6,0.8/.8,1/1.0} {
    \draw[gray!25, line width=0.2pt] (0,\y) -- (6.75,\y);
    \draw (0,\y) -- (-0.06,\y) node[left,font=\small] {\lab};
  }
  \draw[->] (0,0) -- (6.85,0);
  \draw[->] (0,0) -- (0,1.08);
  \node[rotate=90,font=\small] at (-0.62,0.5) {Success rate};
  \node[font=\small\bfseries] at (3.38,1.16) {Clean};

  \foreach \xf/\xo/\fast/\ours in {
    0.52/0.94/0.50/0.55,
    1.82/2.24/0.04/0.18,
    3.12/3.54/0.01/0.71,
    4.42/4.84/0.73/0.97,
    5.72/6.14/0.29/0.62} {
    \path[fastbar] (\xf,0) rectangle ++(0.34,\fast);
    \path[oursbar] (\xo,0) rectangle ++(0.34,\ours);
    \pgfmathsetmacro{\xfc}{\xf+0.09}
    \pgfmathsetmacro{\xoc}{\xo+0.25}
    \pgfmathsetmacro{\yf}{\fast+0.025}
    \pgfmathsetmacro{\yo}{\ours+0.025}
    \node[font=\small,anchor=south] at (\xfc,\yf) {\fast};
    \node[font=\small,anchor=south] at (\xoc,\yo) {\ours};
  }
  \node[align=center,font=\small,anchor=north,inner sep=1pt] at (0.86,-0.04) {Click\\alarm};
  \node[align=center,font=\small,anchor=north,inner sep=1pt] at (2.16,-0.04) {Open\\laptop};
  \node[align=center,font=\small,anchor=north,inner sep=1pt] at (3.46,-0.04) {Bread\\basket};
  \node[align=center,font=\small,anchor=north,inner sep=1pt] at (4.76,-0.04) {Container\\plate};
  \node[align=center,font=\small,anchor=north,inner sep=1pt] at (6.06,-0.04) {Object\\stand};
\end{scope}
\begin{scope}[xshift=8.2cm]
  \foreach \y/\lab in {0/0,0.2/.2,0.4/.4,0.6/.6,0.8/.8,1/1.0} {
    \draw[gray!25, line width=0.2pt] (0,\y) -- (6.75,\y);
    \draw (0,\y) -- (-0.06,\y) node[left,font=\small] {\lab};
  }
  \draw[->] (0,0) -- (6.85,0);
  \draw[->] (0,0) -- (0,1.08);
  \node[rotate=90,font=\small] at (-0.62,0.5) {Success rate};
  \node[font=\small\bfseries] at (3.38,1.16) {Randomized};

  \foreach \xf/\xo/\fast/\ours in {
    0.52/0.94/0.44/0.81,
    1.82/2.24/0.03/0.14,
    3.12/3.54/0.00/0.66,
    4.42/4.84/0.61/0.90,
    5.72/6.14/0.34/0.63} {
    \path[fastbar] (\xf,0) rectangle ++(0.34,\fast);
    \path[oursbar] (\xo,0) rectangle ++(0.34,\ours);
    \pgfmathsetmacro{\xfc}{\xf+0.09}
    \pgfmathsetmacro{\xoc}{\xo+0.25}
    \pgfmathsetmacro{\yf}{\fast+0.025}
    \pgfmathsetmacro{\yo}{\ours+0.025}
    \node[font=\small,anchor=south] at (\xfc,\yf) {\fast};
    \node[font=\small,anchor=south] at (\xoc,\yo) {\ours};
  }
  \node[align=center,font=\small,anchor=north,inner sep=1pt] at (0.86,-0.04) {Click\\alarm};
  \node[align=center,font=\small,anchor=north,inner sep=1pt] at (2.16,-0.04) {Open\\laptop};
  \node[align=center,font=\small,anchor=north,inner sep=1pt] at (3.46,-0.04) {Bread\\basket};
  \node[align=center,font=\small,anchor=north,inner sep=1pt] at (4.76,-0.04) {Container\\plate};
  \node[align=center,font=\small,anchor=north,inner sep=1pt] at (6.06,-0.04) {Object\\stand};
\end{scope}
\end{tikzpicture}

\begin{tikzpicture}[x=1cm,y=1cm]
  \path[fastbar] (0,0) rectangle (0.35,0.18);
  \node[right,font=\small] at (0.42,0.09) {Fast-WAM};
  \path[oursbar] (2.2,0) rectangle (2.55,0.18);
  \node[right,font=\small] at (2.62,0.09) {Vid2WAM(Ours)};
\end{tikzpicture}
\caption{Success rates on representative novel RoboTwin tasks. Vid2WAM consistently outperforms Fast-WAM under clean and randomized evaluation.}
\label{fig:robotwin_novel_gains}
\end{figure*}
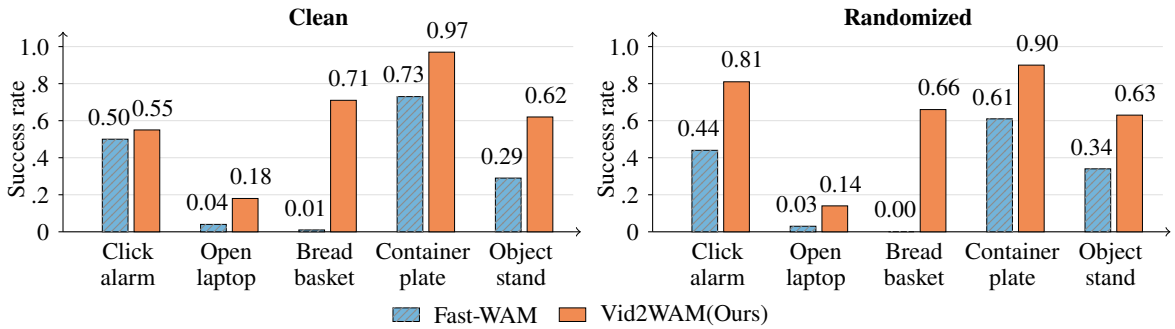

\paragraph{Real-World Setup.}
Our real-world platform consists of two AgileX Piper arms, one overhead Intel
RealSense D435 camera, and two wrist-mounted D435 cameras. The policy takes all
three views, arm joint angles and gripper states as input and predicts
joint-action chunks for both arms. We evaluate six seen tasks and three
held-out novel tasks spanning household and scientific manipulation.
Each seen task provides 60 real trajectories collected with variations in
object placement, lighting, and distractors. Each held-out task
provides no real trajectory and only two true initial observations, one on clean environment with only task-specific objects and one on randomized setup with distractors.
We then use GPT-Image-2 to generate 60 diverse initial-state images from the real
observations for conditioning the video teacher. Because the generated images
lack multi-view alignment, pseudo supervision for held-out tasks is generated
from the overhead view only; the complete policy retains multi-view input when
real multi-view observations are available. Detailed descriptions of real-world setup are provided in Appendix C.

\paragraph{Evaluation Regimes.}
Standard manipulation benchmarks typically train policies with expert trajectories from every evaluation task and test them under limited variations in object configurations. Such protocols primarily measure in-distribution execution and may conflate task generalization with interpolation around task-specific demonstrations. To more directly evaluate data efficiency and novel-task generalization, we consider two complementary regimes. In the \textbf{low-data regime}, each task provides only 10 expert trajectories, with Vid2WAM receiving an equal number of offline teacher-generated trajectories. In the \textbf{novel-task regime}, we withhold all expert trajectories from selected target tasks: baselines are directly transferred from the remaining tasks, whereas Vid2WAM uses only target-task instructions and initial observations to generate pseudo supervision. Neither the video teacher nor the IDM is trained on target-task trajectories. These regimes respectively evaluate policy learning under scarce real demonstrations and adaptation to novel tasks without target-task expert supervision.

\paragraph{Baselines.}
We compare with $\pi_{0.5}$~\cite{physicalintelligence2025pi05}, Motus~\cite{bi2025motusunifiedlatentaction}, and Fast-WAM~\cite{yuan2026fastwam}. $\pi_{0.5}$ and Motus assess whether broad
robot-data pretraining alone enables direct transfer to the evaluated tasks.
Fast-WAM serves as the closest architectural baseline, isolating the effect of
our teacher-derived future supervision and source-aware adaptation. Under each
evaluation regime, all methods receive the same downstream real
action-labeled demonstrations.

\paragraph{Implementation Details.}
We initialize our video teacher from the LVP checkpoint~\citep{chen2025large},
a Wan2.1-14B model~\citep{wan2025wan} pretrained on robot manipulation data,
and further fine-tune it on our training set. Generated frames are encoded by the student VAE so that the teacher-generated targets and student predictions lie in the same latent space. The IDM uses a ResNet-50 backbone~\cite{he2016deep} with action and proprioception prediction heads.
% The source-aware residual adapters use a bottleneck width of $D_r=128$ and GELU activations. We set the loss weights to $\lambda_{\mathrm{real}}=1.0$ and $\lambda_{\mathrm{pseudo}}=0.25$. 
To reduce the influence of long-horizon generation errors, pseudo supervision in simulation experiments is restricted to the first eight seconds of each generated rollout. For real-world novel-task pseudo data, the spatial video loss is computed only over the valid overhead-view region, while pseudo-action supervision remains unchanged. Complete training schedules, IDM validation results, multi-view input layouts, and spatial and temporal masking details are provided in Appendix A\& C.

\subsection{Simulation Results}
\label{sec:simulation_results}

\begin{table}[t!]
\centering
\small
\setlength{\tabcolsep}{3pt}
\begin{tabular}{lcccccc}
\toprule
\multirow{2}{*}{Method}
& \multicolumn{2}{c}{Low-data}
& \multicolumn{2}{c}{Novel regime}
& \multicolumn{2}{c}{Novel subset} \\
\cmidrule(lr){2-3} \cmidrule(lr){4-5} \cmidrule(lr){6-7}
& Clean & Rand. & Clean & Rand. & Clean & Rand. \\
\midrule
$\pi_{0.5}$ & 48.7 & 42.8 & 67.9 & 66.5 & 46.5 & 46.3 \\
Motus       & 49.9 & 43.6 & 75.1 & 76.3 & 48.9 & 51.5 \\
Fast-WAM    & 52.1 & 39.1 & 75.7 & 74.7 & 45.0 & 42.8 \\
Ours     & \textbf{55.4} & \textbf{45.4}
            & \textbf{78.3} & \textbf{78.5}
            & \textbf{54.7} & \textbf{55.3} \\
\bottomrule
\end{tabular}
\caption{Overall success rates (\%) on RoboTwin~2.0. Novel regime averages over
35 seen and 15 unseen tasks, while Novel subset reports the average over 
15 unseen tasks only.}
\label{tab:robotwin_overall}
\end{table}

\begin{figure*}[!t]
\centering
\includegraphics[width=\textwidth]{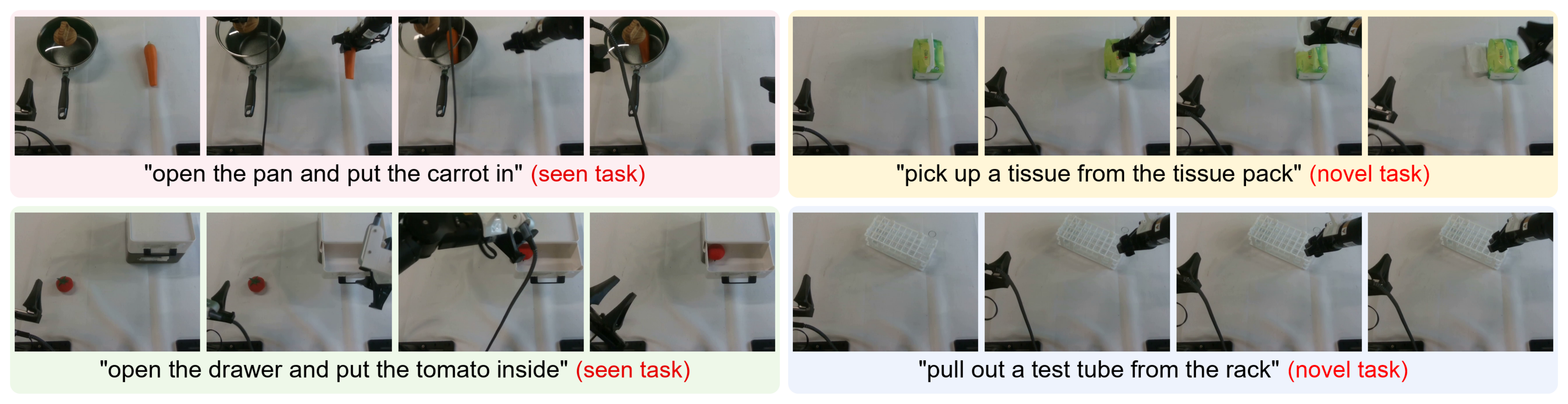}
\caption{Vid2WAM real-world executions for representative seen and novel tasks. Frames progress left to right.}
\label{fig:realworld_rollouts}
\end{figure*}

\label{sec:liberoplus}
\begin{table*}[!t]
\centering
\small
\begin{tabular*}{0.94\textwidth}{@{\extracolsep{\fill}}ll|cccc|ccccccc|c@{}}
\toprule
\multirow{2}{*}{Regime} & \multirow{2}{*}{Model}
& \multicolumn{4}{c|}{Suite}
& \multicolumn{7}{c|}{Perturbation}
& \multirow{2}{*}{Overall} \\
\cmidrule(lr){3-6} \cmidrule(lr){7-13}
& & Spatial & Object & Goal & Long
& Cam. & Robot & Lang. & Light & BG & Noise & Layout & \\
\midrule
\multirow{3}{*}{Low-data}
& Motus & 34.2 & 35.4 & \textbf{29.4} & 12.4 & 2.5 & 29.5 & 53.3 & 38.1 & 31.0 & 6.0 & 38.2 & 27.9 \\
& Fast-WAM & 39.8 & 54.0 & 14.6 & \textbf{25.8} & \textbf{8.5} & 22.5 & 49.5 & 57.8 & 32.1 & 23.3 & 45.6 & 33.6 \\
& Ours & \textbf{50.0} & \textbf{55.0} & 27.8 & 23.0 & 8.2 & \textbf{31.2} & \textbf{57.9} & \textbf{71.6} & \textbf{36.8} & \textbf{23.7} & \textbf{48.8} & \textbf{39.0} \\
\midrule
\multirow{3}{*}{Novel}
& Motus & 52.2 & 44.0 & 31.8 & \textbf{39.6} & 4.7 & \textbf{51.6} & 66.3 & 63.8 & \textbf{46.2} & 18.7 & 47.7 & 41.9 \\
& Fast-WAM & 48.0 & 59.6 & 37.0 & 38.6 & 24.1 & 42.5 & 64.2 & 67.2 & 39.0 & 34.6 & \textbf{52.7} & 45.8 \\
& Ours & \textbf{52.4} & \textbf{60.2} & \textbf{41.4} & 38.6 & \textbf{28.5} & 42.8 & \textbf{67.7} & \textbf{67.5} & 39.7 & \textbf{42.4} & 51.6 & \textbf{48.1} \\
\bottomrule
\end{tabular*}%
\caption{Average task success rates (\%) on LIBERO-Plus under the low-data and novel regimes, evaluated across four suites and seven perturbation types.}
\label{tab:libero_plus_results1}
\label{tab:libero_plus_results}
\end{table*}

% \label{sec:liberoplus}
% \begin{table*}[!t]
% \centering
% \small
% \setlength{\tabcolsep}{4.5pt}
% \begin{tabular}{ll|cccc|ccccccc|c}
% \toprule
% \multirow{2}{*}{Regime} & \multirow{2}{*}{Model}
% & \multicolumn{4}{c|}{Suite}
% & \multicolumn{7}{c|}{Perturbation}
% & \multirow{2}{*}{Overall} \\
% \cmidrule(lr){3-6} \cmidrule(lr){7-13}
% & & Spatial & Object & Goal & Long
% & Cam. & Robot & Lang. & Light & BG & Noise & Layout & \\
% \midrule
% \multirow{3}{*}{Low-data}
% & Motus & 34.2 & 35.4 & \textbf{29.4} & 12.4 & 2.5 & 29.5 & 53.3 & 38.1 & 31.0 & 6.0 & 38.2 & 27.9 \\
% & Fast-WAM & 39.8 & 54.0 & 14.6 & \textbf{25.8} & \textbf{8.5} & 22.5 & 49.5 & 57.8 & 32.1 & 23.3 & 45.6 & 33.6 \\
% & Ours & \textbf{50.0} & \textbf{55.0} & 27.8 & 23.0 & 8.2 & \textbf{31.2} & \textbf{57.9} & \textbf{71.6} & \textbf{36.8} & \textbf{23.7} & \textbf{48.8} & \textbf{39.0} \\
% \midrule
% \multirow{3}{*}{Novel}
% & Motus & 52.2 & 44.0 & 31.8 & \textbf{39.6} & 4.7 & \textbf{51.6} & 66.3 & 63.8 & \textbf{46.2} & 18.7 & 47.7 & 41.9 \\
% & Fast-WAM & 48.0 & 59.6 & 37.0 & 38.6 & 24.1 & 42.5 & 64.2 & 67.2 & 39.0 & 34.6 & \textbf{52.7} & 45.8 \\
% & Ours & \textbf{52.4} & \textbf{60.2} & \textbf{41.4} & 38.6 & \textbf{28.5} & 42.8 & \textbf{67.7} & \textbf{67.5} & 39.7 & \textbf{42.4} & 51.6 & \textbf{48.1} \\
% \bottomrule
% \end{tabular}%
% \caption{Average task success rates (\%) on LIBERO-Plus under the low-data and novel regimes, evaluated across four suites and seven perturbation types.}
% \label{tab:libero_plus_results1}
% \label{tab:libero_plus_results}
% \end{table*}

\begin{table}[t]
\centering
\small
\setlength{\tabcolsep}{5pt}
\begin{tabular}{@{}lccccc@{}}
\toprule
Method/Regime & Spatial & Object & Goal & Long & \textbf{Avg.} \\
\midrule
\multicolumn{6}{l}{\textbf{Low-data}} \\
\midrule

$\pi_{0.5}$
& 91.2 & 95.6 & 84.4 & 76.4 & 86.9 \\
Motus
& \textbf{93.0} & 94.4 & 85.0 & 78.4 & 87.7 \\
Fast-WAM
& 89.6 & 97.8 & 82.8 & 78.8 & 87.3 \\
Ours
& \textbf{93.0} & \textbf{98.6} & \textbf{87.6} & \textbf{79.4} & \textbf{89.7} \\
\midrule
\multicolumn{6}{l}{\textbf{Novel}} \\
\midrule

$\pi_{0.5}$
& 77.2 & \textbf{80.2} & 78.4 & \textbf{75.8} & 77.9 \\
Motus
& 77.4 & 72.2 & 68.8 & 69.8 & 72.1 \\
Fast-WAM
& 76.8 & 79.0 & 77.8 & 73.0 & 76.7 \\
Ours
& \textbf{79.4} & 79.4 & \textbf{79.2} & 75.2 & \textbf{78.3} \\
\bottomrule
\end{tabular}
\caption{Average task success rates (\%) across the four LIBERO suites under the low-data and novel regimes.}
\label{tab:libero}
\label{tab:libero1}
\end{table}

\paragraph{RoboTwin 2.0.}
Table~\ref{tab:robotwin_overall} reports results under clean and randomized
conditions on RoboTwin 2.0, including averages over all 50 tasks and over the 15 novel tasks in the novel regime. The results reveal that both pretrained VLA policies and WAMs struggle to transfer to novel tasks, with all baselines dropping substantially on the 15-task novel subset. This degradation is particularly pronounced for Fast-WAM, whose success rate decreases from 75.7\% to 45.0\% under clean evaluation and from 74.7\% to 42.8\% under randomization. By distilling the pretrained video teacher's predictive prior, Vid2WAM improves novel-subset success to 54.7\% and 55.3\%, outperforming Fast-WAM by 9.7 and 12.5 points. Vid2WAM also achieves the best performance in the low-data regime, demonstrating consistent benefits for both novel-task transfer and learning from limited real demonstrations.

Figure~\ref{fig:robotwin_novel_gains} presents representative novel tasks for
which Vid2WAM clearly improves over Fast-WAM. The teacher-generated future
trajectories provide useful information about task completion and guide the
policy toward the intended outcome. Figure~\ref{fig:qualitative_video_predictions}
further compares the predicted future videos produced from identical initial
observations and language instructions. Fast-WAM often stops after an
intermediate interaction or loses visual coherence, while Vid2WAM continues
toward the requested placement. The difference becomes most visible after the
policy has completed an opening or grasping step, suggesting that the distilled
future information supports longer-horizon task progression.

\paragraph{LIBERO and LIBERO-Plus.}
Table~\ref{tab:libero} reports results on LIBERO under both the low-data and
novel regimes. In the low-data regime, Vid2WAM achieves an average success rate
of $89.7\%$, obtaining the best or tied-best result on every suite. In the
novel regime, it also achieves the highest average success rate among the four
evaluated methods. 
These results show that Vid2WAM improves performance both
when expert demonstrations are limited and when no real trajectories are
available for the target tasks.

Table~\ref{tab:libero_plus_results} further evaluates robustness under the
perturbations introduced by LIBERO-Plus.
Vid2WAM achieves the highest overall performance on both regimes, with particularly
clear gains under robot-state, language, and lighting perturbations on the
low-data regime. We further evaluate whether the novel regime results depend
on the selected target tasks. We consider three LIBERO or LIBERO-Plus
splits that withhold novel task IDs $0/1$, $3/6$, or $8/9$ from each suite, where the $8/9$ split corresponds to the settings in Table~\ref{tab:libero_plus_results} and Table~\ref{tab:libero}. Results in Appendix B show that Vid2WAM always achieves a higher average success rate than Fast-WAM on all three splits in both benchmarks.
Results in Appendix B also evaluate only the novel tasks from these LIBERO
and LIBERO-Plus splits, and Vid2WAM again outperforms Fast-WAM on every split,
showing that the aggregate improvements result from better adaptation to novel
tasks rather than the selection of some specific tasks.

\subsection{Real-World Experiments}

We evaluate real-world performance on nine contact-rich manipulation tasks,
covering both seen and novel settings. Seen tasks include
\textit{Click Bell}, \textit{Item Handover}, 
\textit{Carry Basket}, \textit{Open Pan Place}, 
\textit{Open Drawer Place} and \textit{Insert Test Tube}. 
They use real trajectories during training. Novel tasks
include \textit{Pick Test Tube}, \textit{Take Tissue} and 
\textit{Tomato Basket}. They are designed to test behavior-level or object-level novelty,
requiring novel manipulation actions or new objects that are absent from the real expert trajectories. These tasks use initial observations
(true and synthetic) and teacher-generated supervision rather than real demonstrations. Detailed real-world task descriptions are provided in Appendix C.

Table~\ref{tab:realworld_results} reports success rates on 20 real-robot trials for all models on each task. Among all tasks, Vid2WAM achieves the best or competitive performance across the four models.
Figure~\ref{fig:realworld_rollouts} further visualizes representative real-world
executions of Vid2WAM on seen and novel tasks. In the novel task, the robot can extract a deformable tissue or pull out a transparent test tube despite having no real
action-labeled trajectories. 

\begin{table}[t]
\centering
\small
\setlength{\tabcolsep}{3pt}
\begin{tabular}{llcccc}
\toprule
Regime & Task & $\pi_{0.5}$ & Motus & Fast-WAM & Ours \\
\midrule
\multirow{6}{*}{Seen}
& Click Bell & 30\% & \textbf{70\%} & 20\% & 65\% \\
& Item Handover & 25\% & \textbf{60\%} & 25\% & \textbf{60\%} \\
& Carry Basket & 45\% & 45\% & 20\% & \textbf{50\%} \\
& Open Pan Place & 5\% & \textbf{40\%} & 10\% & \textbf{40\%} \\
& Open Drawer Place & 60\% & 55\% & 30\% & \textbf{65\%} \\
& Insert Test Tube & 15\% & 0\% & 10\% & \textbf{25\%} \\
\midrule
\multirow{3}{*}{Novel}
& Pick Test Tube & 0\% & 10\% & 5\% & \textbf{30\%} \\
& Take Tissue & 0\% & 0\% & 5\% & \textbf{15\%} \\
& Tomato Basket & 0\% & 20\% & 25\% & \textbf{30\%} \\
\bottomrule
\end{tabular}%
\caption{Real-world bimanual task success rates.}
\label{tab:realworld_results}
\end{table}

\subsection{Ablation Studies}
\label{sec:ablation}
Table~\ref{tab:libero_Ablation} evaluates on different variants of our Vid2WAM under LIBERO low-data regime.
\emph{Dual Action DiT} variant uses separate action DiTs for ground-truth and
teacher-generated samples while sharing the video DiT; \emph{No Adapter} mixes
both sources in shared action and video DiTs.
% The full Vid2WAM instead retains a shared
% action DiT and uses source-specific residual adapters to separate the two action
% domains. 
\emph{Pseudo Action Only} retains Vid2WAM's synthetic rollouts and adapters but
removes the latent-video objective, while
\emph{Future Latent Only}
removes pseudo-action supervision and uses only future latents for distillation.

Results show that among the dual-channel variants, independent action DiTs outperform fully
shared training, while the full adapter design (Ours) performs best on average. 
\emph{Future Latent Only} outperforms \emph{Pseudo Action Only} by $1.8$ points on average and
Fast-WAM by $0.9$ points, showing that future latents provide a useful guidance signal. Its lower average than our full Vid2WAM
further suggests that pseudo actions provide complementary supervision. 

% Vid2WAM's
% higher average than \emph{No Adapter} provides empirical support that
% source-dependent residual corrections mitigate cross-source interference while
% retaining useful action-level sharing.

We additionally evaluate an online \emph{Teacher Policy} that composes the video
teacher and IDM at inference time without student distillation.
% the teacher
% generates a rollout, and the IDM recovers the action chunk for execution. 
It underperforms Fast-WAM
and all distilled variants, showing that the gains come from offline future
supervision combined with ground-truth action anchoring rather than a stronger
online teacher controller.
This result is also consistent with generation and IDM errors accumulating
under task execution. Together with the \emph{No Adapter} ablation, it
suggests that source-dependent residual corrections can mitigate the influence
of noisy pseudo-actions.

\begin{table}[t]
\centering
\small
\setlength{\tabcolsep}{3.5pt}
\begin{tabular}{lccccc}
\toprule
Method & Spatial & Object & Goal & Long & \textbf{Avg.} \\
\midrule
Fast-WAM & 89.6 & 97.8 & 82.8 & 78.8 & 87.3 \\
Ours & 93.0 & \textbf{98.6} & \textbf{87.6} & 79.4 & \textbf{89.7} \\
Dual Action DiT & 92.2 & 97.8 & 85.8 & \textbf{80.6} & 89.1 \\
No Adapter & \textbf{93.8} & 97.6 & 84.2 & 78.4 & 88.5 \\
Pseudo Action Only & 90.4 & 96.0 & 85.0 & 74.0 & 86.4 \\
Future Latent Only & 92.8 & 98.2 & 86.4 & 75.2 & 88.2 \\
\midrule
Teacher Policy & 76.0 & 85.2 & 65.6 & 58.2 & 71.3 \\
\bottomrule
\end{tabular}%
\caption{Ablation study on LIBERO benchmark (low-data regime).}
\label{tab:libero_Ablation}
\end{table}

\begin{table}[t]
\centering
\small
\setlength{\tabcolsep}{3.5pt}
\begin{tabular}{lrrrrr}
\toprule
Method & Params & Mean & Std. & Min & Max \\
\midrule
Fast-WAM & 5B &209.1 & 4.7 & 204.8 & 220.2 \\
Ours & 5B & 212.5 & 11.7 & 199.0 & 227.8 \\
Teacher Policy & 14B & 4894.5 & 106.6 & 4785.8 & 4998.9 \\
$\pi_{0.5}$ & 3.3B & 76.7 & 0.4 & 76.2 & 77.6 \\
Motus & 8B & 2414.9 & 22.8 & 2380.0 & 2456.3 \\
\bottomrule
\end{tabular}%
\caption{Inference latency (ms) on a single RTX 4090 GPU.}
\label{tab:latency}
\end{table}

% All distillation variants improve the average over Fast-WAM, while independent action DiTs outperform fully shared
% training, which suggests that separating heterogeneous action supervision can
% reduce interference of noise from pseudo data. The full adapter design performs best overall, which shows lightweight
% source-specific adapters provide a better balance between useful
% action-level sharing and source-dependent separation.

\subsection{Inference Latency Analysis}
\label{sec:latency}
Table~\ref{tab:latency} reports chunk-level inference latency on a single RTX
4090 GPU, excluding cold-start costs. Results show that our Vid2WAM approximately matches Fast-WAM in latency, and is substantially faster than Motus and
the Teacher Policy, though slower than $\pi_{0.5}$.

%% file: sections/5_conclusion.tex
% \section{Limitations}
% \label{sec:limits}
% TODO: latency against pi05/vla;pi05 libero-plus;framework on other wams

\section{Conclusion}

We present Vid2WAM, a framework for transferring the predictive diffusion priors of a large video model into a compact world-action policy. By using teacher-generated futures and IDM-inferred actions as offline supervision, Vid2WAM enables the student to acquire behavioral knowledge beyond the available expert trajectories, while source-aware adaptation limits the impact of noisy synthetic action labels. Across simulation and real-world manipulation, Vid2WAM improves data efficiency and generalization to novel tasks without introducing teacher inference at deployment. Our findings support a broader paradigm in which large pretrained models serve as offline knowledge sources for efficient robot policies. Video generators, VLMs, and other multimodal foundation models could supervise compact policies through predicted subgoals, future representations, or action-conditioned outcomes, offering a scalable path for transferring their predictive capabilities to efficient robot policies.

%% file: supp/sections/teacher_idm.tex
\section{Implementation Details}
  For all models including video teacher, IDM and Vid2WAM, on LIBERO and LIBERO-Plus, the head and wrist camera views are resized to
  $224 \times 224$ and concatenated horizontally into a $224 \times 448$ frame.
  For RoboTwin 2.0, the top head camera
  is resized to $256 \times 320$, while the two wrist views are resized to
  $128 \times 160$, concatenated horizontally, and placed below to
  form a $384 \times 320$ canvas. 
For real-world experiments, the video teacher generates overhead-view rollouts at its native $480 \times 640$ resolution. For student training, these frames are resized to occupy the upper $224 \times 448$ region of a $448 \times 448$ composite, while the two unavailable $224 \times 224$ wrist-view slots in the lower half are black-padded. For real multi-view samples, the true wrist views occupy these lower slots.

 \subsection{Teacher and Inverse Dynamics Model Training}
  \label{app:teacher_idm_training}

  \paragraph{Video teacher training.}
  The teacher is an image-to-video diffusion model initialized from the Large Video Planner checkpoint~\cite{chen2025large}, which adopts the Wan2.1-I2V-14B architecture~\cite{wan2025wan}. We fine-tune the diffusion transformer while keeping the UMT5 text encoder, CLIP image encoder, and video VAE frozen. All video teachers are trained for around 500 epochs. Each training sample consists of 49 video frames, which are encoded into 16-channel latent representations using the frozen Wan VAE.

  Given a clean video latent \(z\), Gaussian noise
  \(\epsilon\sim\mathcal{N}(0,I)\), and \(t\sim\mathcal{U}(0,1)\), we apply the
  time shift
  \begin{equation}
  \tau=\frac{s t}{1+(s-1)t},
  \qquad s=3,
  \end{equation}
  and construct
  \begin{equation}
  z_{\tau}=(1-\tau)z+\tau\epsilon.
  \end{equation}
  The teacher predicts the flow target \(v^\star=\epsilon-z\) and is optimized
  with
  \begin{equation}
  \mathcal{L}_{\mathrm{teacher}}
  =
  \mathbb{E}_{z,\epsilon,t}
  \left[
  \left\|
  v_{\Theta}(z_{\tau},\tau,c)-(\epsilon-z)
  \right\|_2^2
  \right],
  \end{equation}
  where \(c\) contains the language embedding and first-frame image condition.
  We additionally use diffusion forcing~\cite{chen2024diffusion} during training. In particular, one or
  two latent history frames are sampled with probabilities \(0.9\) and \(0.1\),
  respectively, and the history is kept clean with probability \(0.7\).

  \begin{table}[t]
  \centering
  \footnotesize
  \renewcommand{\arraystretch}{0.96}
  \begin{tabularx}{\linewidth}{lY}
  \toprule
  Hyperparameter & Value \\
  \midrule
  Initialization
    & Wan2.1-I2V-14B-480P with LVP 14B weights \\
  Layers
    & 40 \\
  Hidden dimension
    & 5120 \\
  Feed-forward dimension
    & 13824 \\
  Attention heads
    & 40 \\
  Training frames
    & 49 \\
  VAE latent channels
    & 16 \\
  VAE temporal/spatial stride
    & \(4\times 8\times 8\) \\
  % Training objective
  %   & Flow-matching mean-squared error \\
  % Flow target
  %   & \(\epsilon-z\) \\
  Continuous-time shift
    & 3.0 \\
  Training-time scale
    & 1000 \\
  % Diffusion forcing
  %   & \texttt{rand\_history} \\
  % History length distribution
  %   & 1 or 2 latent frames with probabilities \(0.9/0.1\) \\
  % Clean-history probability
  %   & 0.7 \\
  Optimizer
    & AdamW \\
  Learning rate
    & \(8\times10^{-6}\) \\
  AdamW betas
    & \((0.9,0.95)\) \\
  Weight decay
    & \(5\times10^{-2}\) \\
  Warm-up steps
    & 100 \\
  Batch size
    & 4 \\
  Gradient accumulation
    & 4 \\
  Numerical precision
    & BF16 \\
  Maximum text length
    & 512 \\
  \bottomrule
  \end{tabularx}
  \normalsize
  \caption{Shared hyperparameters used to fine-tune the video teacher.}
  \label{tab:video_teacher_hparams}
  \end{table}

  \paragraph{Inverse dynamics model.}
  The inverse dynamics model (IDM) maps a short video window to the action at
  its center frame. Given three observations separated by a domain-dependent
  stride \(\delta\), it predicts
  \begin{equation}
  \widehat{a}_t
  =
  q_{\psi}\left(
  I_{t-\delta},I_t,I_{t+\delta}
  \right).
  \end{equation}
  We use an ImageNet-pretrained ResNet-50~\cite{he2016deep} whose first convolution is expanded
  from three to nine input channels to jointly process the three RGB frames.
  The pretrained convolutional filters are repeated across the temporal
  dimension and divided by three. The final ResNet feature map is passed
  through a learnable spatial-softmax layer and a lightweight MLP action head.

  The action mean \(\mu_a\) and standard deviation \(\sigma_a\) are computed
  from the training split. The IDM is trained with a Smooth L1 loss in the
  normalized action space:
  \begin{equation}
  \mathcal{L}_{\mathrm{IDM}}
  =
  \mathbb{E}
  \left[
  \operatorname{SmoothL1}
  \left(
  \frac{\widehat{a}_t-\mu_a}{\sigma_a},
  \frac{a_t-\mu_a}{\sigma_a}
  \right)
  \right].
  \end{equation}
  The prediction is converted back to the original action space during
  inference.

  \begin{table}[t]
  \centering
  \small
  \begin{tabularx}{\linewidth}{lY}
  \toprule
  Hyperparameter & Value \\
  \midrule
  Visual backbone
    & ResNet-50 \\
  Number of input frames
    & 3 \\
  Input channels
    & 9 \\
  Feature pooling
    & Spatial softmax \\
  Hidden activation
    & GELU \\
  Dropout
    & 0.1 \\
  Training loss
    & Smooth L1  \\
  Optimizer
    & AdamW \\
  Learning rate
    & \(3\times10^{-4}\) \\
  AdamW betas
    & \((0.9,0.999)\) \\
  Weight decay
    & \(10^{-2}\) \\
  Training iterations
    & 200,000 \\
  Batch size
    & 32 \\
  Learning-rate schedule
    & Cosine \\
  Warm-up iterations
    & 5,000 \\
  Final learning-rate ratio
    & 0 \\
  Numerical precision
    & BF16 \\

  % Color augmentation
  %   & Color jitter with brightness \(0.8\)--\(1.2\),
  %     contrast \(0.7\)--\(1.3\), saturation \(0.5\)--\(1.5\),
  %     and hue \(0.05\) \\
  \bottomrule
  \end{tabularx}
  \normalsize
  \caption{Shared training hyperparameters for the action IDM.}
  \label{tab:idm_hparams}
  \end{table}

  \begin{table}[t]
  \centering
  \small
  \setlength{\tabcolsep}{3pt}
  \resizebox{\columnwidth}{!}{%
  \begin{tabular}{lccccccc}
  \toprule
  Domain & Resolution & FPS & Frames  &  Steps  & Language CFG & History CFG\\
  \midrule
  LIBERO
    & \(224\times448\) & 15 & 49 & 40 & 3.0 & 1.0\\
  RoboTwin
    & \(384\times320\) & 15 & 49 & 40 & 3.0 & 1.0\\
  Real-world
    & \(480\times640\) & 10 & 49 & 40 & 3.0 & 2.0\\
  \bottomrule
  \end{tabular}
  }
  \normalsize
  \caption{Domain-specific video-teacher rollout configurations used for constructing the pseudo-data buffers.}
  \label{tab:video_teacher_domains}
  \end{table}

\paragraph{IDM training and validation.}
For LIBERO/LIBERO-Plus, RoboTwin and real-world, we train separate ResNet-50 IDMs for
action and proprioceptive-state targets. Each IDM is trained for 200,000 steps using compatible action-labeled trajectories; withheld-task
trajectories are excluded. The frozen IDMs then label pseudo-actions from teacher-generated videos.

  For validation, we randomly partition the frame-level samples with size $N$. The validation-set size is
  \begin{equation}
  N_{\mathrm{val}}
  =
  \min\left(
  \left\lfloor 0.05N \right\rfloor,
  2000
  \right),
  \end{equation}
  with at least one validation sample. Target normalization statistics are
  computed exclusively from the resulting training split. We evaluate the IDM
  every 5,000 optimization steps and retain the checkpoint with the lowest
  validation Smooth L1 loss. We further define validation accuracy as the proportion of samples for which all predicted dimensions fall within their corresponding absolute-error tolerances:
  \begin{equation}
  \mathrm{ValAcc}
  =
  \frac{1}{N}
  \sum_{i=1}^{N}
  \mathbb{I}
  \left(
  \bigcap_{j=1}^{D}
  \left\{
  \left|\hat{y}_{i,j}-y_{i,j}\right| < \tau_j
  \right\}
  \right),
  \end{equation}
  where $N$ is the number of validation samples, $D$ is the target
  dimensionality, $\hat{y}_{i,j}$ and $y_{i,j}$ denote the prediction and
  ground-truth value for dimension $j$ of sample $i$ respectively, and
  $\tau_j=0.05$ for every dimension. Consequently, a sample is counted as
  correct only if every predicted dimension satisfies the tolerance. The rules are identical for action and state IDM training.

Table~\ref{tab:idm_results} reports validation results for all IDMs, with
all models achieving higher than 80\% accuracy, especially both real-world heads exceeding 98\% accuracy.

\begin{table}[t]
  \centering
  \small
  \setlength{\tabcolsep}{3pt}
  \resizebox{\columnwidth}{!}{%
  \begin{tabular}{llccc}
  \toprule
  Dataset & IDM Target & Val Loss $\downarrow$ & Val Acc $\uparrow$ & Val L1 $\downarrow$ \\
  \midrule
  LIBERO / LIBERO-Plus & Action & 0.0054 & 85.8\%  & 0.0112 \\
  LIBERO / LIBERO-Plus & State  & 0.0000 & 100.0\% & 0.0020 \\
  RoboTwin             & Action & 0.0009 & 81.9\%  & 0.0100 \\
  RoboTwin             & State  & 0.0008 & 83.1\%  & 0.0095 \\
  Real-world     & Action & 0.0002 & 98.1\%  & 0.0044 \\
  Real-world      & State  & 0.0002 & 98.6\%  & 0.0043 \\
  \bottomrule
  \end{tabular}
  }
  \normalsize
  \caption{Best validation performance of the trained IDM models.}
  \label{tab:idm_results}
\end{table}

  \paragraph{Pseudo-data construction.}
    After training, the video teacher and IDM are used only as offline data generators. Table~\ref{tab:video_teacher_domains} lists the rollout parameters used
  to construct each distillation buffer.
Rollout generation takes only
the source episode's first visual observation  together with the language instruction, after which the
future observations are synthesized by the video teacher and the pseudo
actions are recovered by the IDM.  

   Because the Wan2.1 teacher VAE uses 16
  latent channels while the Vid2WAM VAE uses 48, teacher rollouts are decoded to
  RGB and re-encoded by the frozen student VAE before storage.

  The 49 frames in Table~\ref{tab:video_teacher_domains} denote one generation
  chunk, not a complete pseudo-rollout. Successive chunks are assembled into
  longer videos before the simulation cutoff and pseudo-data construction.

\begin{figure*}[!t]
  \centering
  \includegraphics[width=0.98\textwidth]{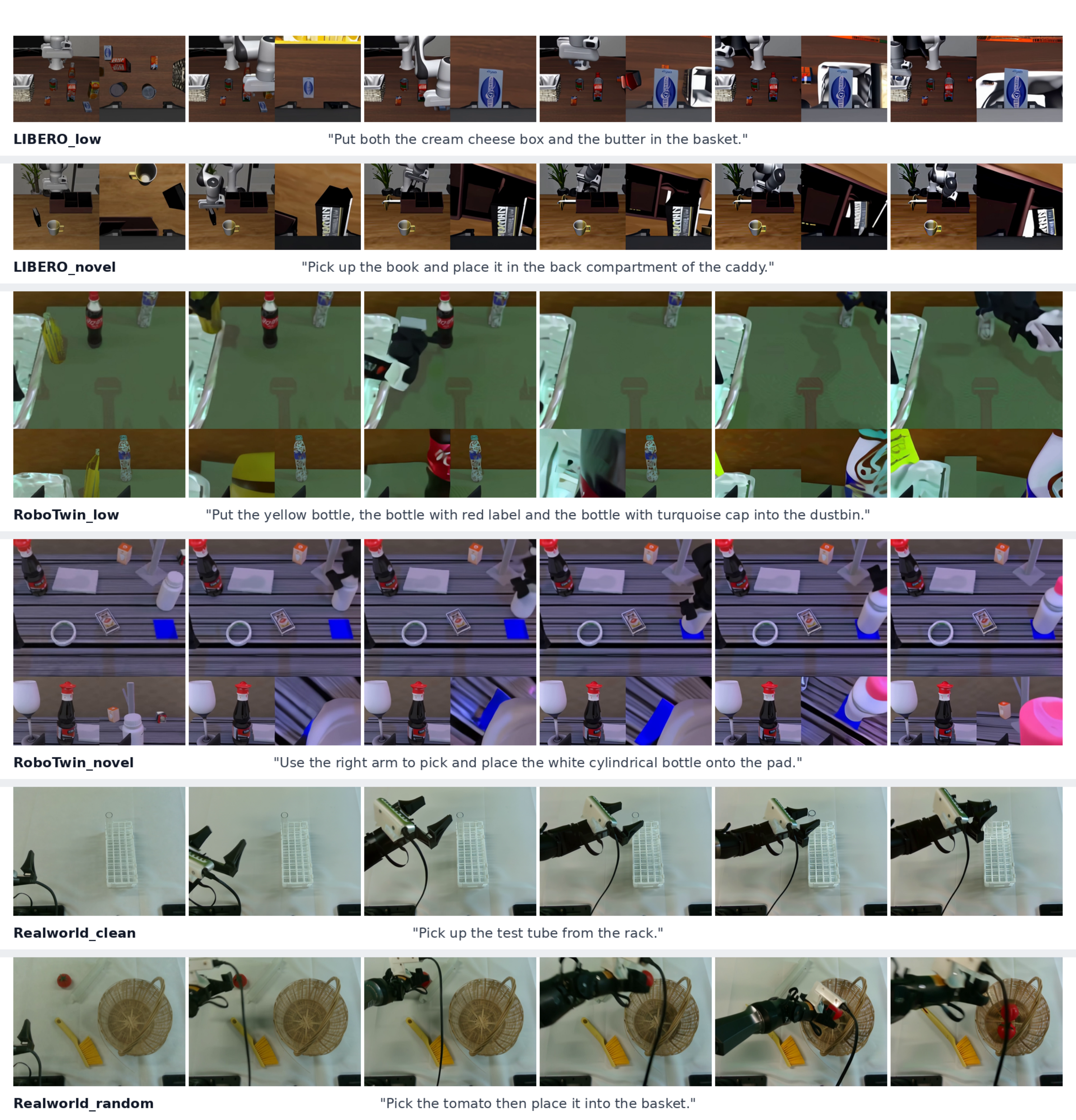}
  \caption{Representative teacher-generated rollouts across LIBERO, RoboTwin,
  and real-world settings. Each row presents six temporally ordered keyframes
  conditioned on the displayed language instruction. The simulation examples
  include low-data and novel-task regimes, while the real-world examples cover
  clean and randomized initial states.}
  \label{fig:teacher_rollout_examples}
\end{figure*}

\paragraph{Teacher-generated rollout examples.}
Figure~\ref{fig:teacher_rollout_examples} shows representative teacher rollouts
across LIBERO, RoboTwin, and real-world settings. The examples cover both
low-data and novel-task regimes, as well as clean and randomized real-world
initial states.

Across the six examples, the teacher predicts task-relevant object motion and
multi-step manipulation progress while maintaining the scene context needed for
the instruction. The novel-task and randomized examples further illustrate
useful future-progress cues for held-out instructions and perturbed initial
states. 

\subsection{Data Budget and Model Exposure}
\label{app:data_budget}

Table~\ref{tab:data_budget} reports the data budget. For all experiment setups, the video teachers and IDMs are fine-tuned on its regime-specific
dataset: the low-data regime uses corresponding expert demonstrations, while the novel regime excludes held-out-task trajectories.

\begin{table*}[!t]
  \centering
  \small
  \setlength{\tabcolsep}{4pt}
  \renewcommand{\arraystretch}{1.06}
  \begin{tabularx}{\textwidth}{@{}lcc>{\raggedright\arraybackslash}Xrcc@{}}
    \toprule
    Benchmark/regime
      & \multicolumn{2}{c}{Tasks}
      & Expert trajectories (seen tasks only)
      & Teacher rollouts
      & \multicolumn{2}{c}{Held-out-task exposure} \\
    \cmidrule(lr){2-3}\cmidrule(l){6-7}
      & Seen & Novel & & & Teacher & IDM \\
    \midrule
    LIBERO low-data
      & 40 & 0 & 400 (10 per task)
       & 400  & / & / \\
    LIBERO novel
      & 32 & 8 & 1,366
       & 312  & No & No \\
    RoboTwin low-data
      & 50 & 0 & 1000 (10 clean $+$ 10 randomized per task)
       & 1000  & / & / \\    
    RoboTwin novel
      & 35 & 15 & 19250 (50 clean $+$500 randomized per seen-task)
       & 8250  & No & No \\
    Real-world
      & 6 & 3 & 360 (60 per seen-task)
       & 180  & No & No \\
    \bottomrule
  \end{tabularx}
  \normalsize
  \caption{All data budgets and exposures. ``Teacher exposure'' and
  ``IDM exposure'' indicate whether the corresponding model was trained on
  trajectories from the evaluated held-out target tasks; this is distinct from using
  a trajectory's first frame to seed a rollout.}
  \label{tab:data_budget}
\end{table*}

%% file: supp/sections/implementation_detail.tex
\subsection{Vid2WAM Training Details}
  \label{app:implementation_details}

  \paragraph{Model and distillation training details.}
  Table~\ref{tab:model_hparams} summarizes the model architecture and training hyperparameters used by Vid2WAM. Each iteration processes one real batch and one pseudo batch before
  a single parameter update. Thus, the reported global batch size of 128 per domain denotes 128 real and 128 pseudo samples per update.

  For LIBERO/LIBERO-Plus and real-world, we trained Vid2WAM for 10 epochs, while for RoboTwin 2.0, we trained for 5 epochs.
  
  \begin{table}[!t]
  \centering
  \footnotesize
  \setlength{\tabcolsep}{3pt}
  \renewcommand{\arraystretch}{0.96}
  \begin{tabularx}{\columnwidth}{@{}>{\raggedright\arraybackslash}p{0.48\columnwidth}Y@{}}
  \toprule
  Hyperparameter & Value \\
  \midrule
  Video backbone initialization & Wan2.2-TI2V-5B \\
  Number of transformer layers & 30 \\
  Number of attention heads & 24 \\
  Attention head dimension & 128 \\
  Video hidden dimension & 3072 \\
  Video FFN dimension & 14336 \\
  Action hidden dimension $D$ & 1024 \\
  Action FFN dimension & 4096 \\
  Text embedding dimension & 4096 \\
  Maximum text length & 128 \\
  Video patch size & $(1,2,2)$ \\
  \midrule
  % Adapter domains & Real and pseudo \\
  % Adapters per domain & Input and output \\
  Bottleneck dimension $D_r$ & 128 \\
  Adapter activation & GELU \\
  Adapter dropout & 0.0 \\
  Adapter residual scale $\alpha$ & 1.0 \\
  Up-projection initialization & Zero \\
  \midrule
  Flow-matching training steps & 1000 \\
  Video noise-schedule shift & 5.0 \\
  Action noise-schedule shift & 5.0 \\
  Balance coefficient $\beta$ & 1.0 \\
  $\lambda_{\mathrm{real}}$ & 1.0 \\
  $\lambda_{\mathrm{pseudo}}$ & 0.25 \\
  \midrule
  Global batch per domain & 128  \\
  Gradient accumulation & 1  \\
  Learning rate
    & LIBERO/RoboTwin: $1\times10^{-4}$; real-world: $5\times10^{-5}$ \\
  LR schedule & Cosine  \\
  Warmup ratio & 5\%  \\
  Minimum LR ratio & 0.01  \\
  Optimizer & AdamW  \\
  AdamW $\beta_1,\beta_2$ & $(0.9,0.95)$  \\
  Weight decay & 0.01 \\
  Maximum gradient norm & 1.0  \\
  Numerical precision & BF16 \\
  \bottomrule
  \end{tabularx}
  \normalsize
  \caption{Model and action-adapter hyperparameters in Vid2WAM.}
  \label{tab:model_hparams}
  \end{table}

\paragraph{Details on ablation variants.}

Details on the Vid2WAM-based ablation variants discussed in the main paper are shown in Table~\ref{tab:ablation_implementation}. All Vid2WAM-based variants share the same teacher and IDM models and corresponding distillation buffers. Training hyperparameters also remain the same.

For the Teacher Policy described in the main paper, we also use the same teacher and IDM models as Vid2WAM across all regimes and benchmarks. This policy uses the teacher to generate task rollouts and the IDM to infer action trajectories for direct execution.

\begin{table*}[!t]
  \centering
  \setlength{\tabcolsep}{4pt}
  \small
  \begin{tabularx}{\textwidth}{@{}>{\raggedright\arraybackslash}p{0.14\textwidth}
      Y>{\raggedright\arraybackslash}p{0.17\textwidth}Y@{}}
    \toprule
    Variant & Structural or objective change & Training-time parameter change & Inference path \\
    \midrule
    Vid2WAM (default)
      & Shared video expert and shared ActionDiT; separate input/output
        adapters for both domains.
      & around $+$1M
      & Shared ActionDiT with the real-domain adapters. \\
    Dual Action DiT
      & Shared video expert, but independent real and pseudo
        Action DiTs.
      & around $+$1B
      & Real ActionDiT only \\
    No Adapter
      & One shared video expert and one shared Action DiT, with no
        domain-specific adapter modules. Both domains update the same path.
      & 0
      & The shared Action DiT path. \\
    Pseudo-Action Only
      & Discard the future latent flow-matching loss.
      & Same as Vid2WAM
      & Same as Vid2WAM \\
    Future-Latent Only
      & Discard the pseudo-action flow-matching loss.
      & Same as Vid2WAM
      & Same as Vid2WAM \\
    \bottomrule
  \end{tabularx}
  \normalsize
  \caption{Exact implementation of the distillation ablations.}
  \label{tab:ablation_implementation}
\end{table*}

%% file: supp/sections/task_selection_sensitivity.tex
\section{Sensitivity to Novel-Task Selection}

The split reported in the main paper withholds task IDs $8/9$ from every LIBERO and LIBERO-Plus
suite; we additionally evaluate $0/1$ and $3/6$. Tables
\ref{tab:libero_different_task} and
\ref{tab:libero_plus_split_sensitivity} report full-benchmark performance,
whereas Tables~\ref{tab:libero_novel_success} and
\ref{tab:libero_plus_novel_success} evaluate only the withheld tasks.

\subsection{Task-ID Mapping}

Task IDs are suite-local, and LIBERO-Plus uses the same mapping as LIBERO.
Tables~\ref{tab:libero_task_mapping} and
\ref{tab:libero_task_mapping_goal_long} list the task-ID mappings for all four
suites. A split withholds corresponding tasks with specified IDs within each suite.

\begin{table*}[!t]
\centering
\small
\setlength{\tabcolsep}{3pt}
\renewcommand{\arraystretch}{1.05}
\begin{tabularx}{\textwidth}{@{}cYcY@{}}
\toprule
ID & LIBERO-Spatial instruction & ID & LIBERO-Object instruction \\
\midrule
0 & pick up the black bowl between the plate and the ramekin and place it on
    the plate
  & 0 & pick up the alphabet soup and place it in the basket \\
1 & pick up the black bowl from table center and place it on the plate
  & 1 & pick up the bbq sauce and place it in the basket \\
2 & pick up the black bowl in the top drawer of the wooden cabinet and place
    it on the plate
  & 2 & pick up the butter and place it in the basket \\
3 & pick up the black bowl next to the cookie box and place it on the plate
  & 3 & pick up the chocolate pudding and place it in the basket \\
4 & pick up the black bowl next to the plate and place it on the plate
  & 4 & pick up the cream cheese and place it in the basket \\
5 & pick up the black bowl next to the ramekin and place it on the plate
  & 5 & pick up the ketchup and place it in the basket \\
6 & pick up the black bowl on the cookie box and place it on the plate
  & 6 & pick up the milk and place it in the basket \\
7 & pick up the black bowl on the ramekin and place it on the plate
  & 7 & pick up the orange juice and place it in the basket \\
8 & pick up the black bowl on the stove and place it on the plate
  & 8 & pick up the salad dressing and place it in the basket \\
9 & pick up the black bowl on the wooden cabinet and place it on the plate
  & 9 & pick up the tomato sauce and place it in the basket \\
\bottomrule
\end{tabularx}
\normalsize
\caption{Suite-local LIBERO-Spatial and LIBERO-Object task-ID
mappings, shared by LIBERO and LIBERO-Plus.}
\label{tab:libero_task_mapping}
\end{table*}

\begin{table*}[!t]
\centering
\small
\setlength{\tabcolsep}{3pt}
\renewcommand{\arraystretch}{1.05}
\begin{tabularx}{\textwidth}{@{}cYcY@{}}
\toprule
ID & LIBERO-Goal instruction & ID & LIBERO-Long instruction \\
\midrule
0 & open the middle drawer of the cabinet
  & 0 & turn on the stove and put the moka pot on it \\
1 & open the top drawer and put the bowl inside
  & 1 & put the black bowl in the bottom drawer of the cabinet and close it \\
2 & push the plate to the front of the stove
  & 2 & put the yellow and white mug in the microwave and close it \\
3 & put the bowl on the plate
  & 3 & put both moka pots on the stove \\
4 & put the bowl on the stove
  & 4 & put both the alphabet soup and the cream cheese box in the basket \\
5 & put the bowl on top of the cabinet
  & 5 & put both the alphabet soup and the tomato sauce in the basket \\
6 & put the cream cheese in the bowl
  & 6 & put both the cream cheese box and the butter in the basket \\
7 & put the wine bottle on the rack
  & 7 & put the white mug on the left plate and put the yellow and white mug on
    the right plate \\
8 & put the wine bottle on top of the cabinet
  & 8 & put the white mug on the plate and put the chocolate pudding to the
    right of the plate \\
9 & turn on the stove
  & 9 & pick up the book and place it in the back compartment of the caddy \\
\bottomrule
\end{tabularx}
\normalsize
\caption{Suite-local LIBERO-Goal and LIBERO-Long task-ID mappings,
shared by LIBERO and LIBERO-Plus.}
\label{tab:libero_task_mapping_goal_long}
\end{table*}

From Table~\ref{tab:libero_different_task} and Table~\ref{tab:libero_plus_split_sensitivity}, Vid2WAM improves the average LIBERO and LIBERO-Plus success rate for all three splits, which shows that Vid2WAM's enhancement over Fast-WAM on the novel regime does not depend on the selected target tasks.

\subsection{Withheld-Task Results}

Tables~\ref{tab:libero_novel_success} and
\ref{tab:libero_plus_novel_success} isolate the withheld target tasks in LIBERO and LIBERO-Plus rather than
averaging over the full benchmark suites.

\begin{table}[H]
\centering
\small
\setlength{\tabcolsep}{2.5pt}
\resizebox{\linewidth}{!}{%
\begin{tabular}{llccccc}
\toprule
Novel IDs & Method & Spatial & Object & Goal & Long & \textbf{Avg.} \\
\midrule
\multirow{2}{*}{$0/1$}
& Fast-WAM & 29.0 & 1.0 & 0.0 & 0.0 & 7.5 \\
& Vid2WAM & \textbf{90.0} & \textbf{63.0} & 0.0 & 0.0 & \textbf{38.3} \\
\midrule
\multirow{2}{*}{$3/6$}
& Fast-WAM & 0.0 & 0.0 & 0.0 & 0.0 & 0.0 \\
& Vid2WAM & \textbf{1.0} & \textbf{26.0} & \textbf{1.0} & 0.0 & \textbf{7.0} \\
\midrule
\multirow{2}{*}{$8/9$}
& Fast-WAM & 0.0 & 0.0 & 0.0 & 0.0 & 0.0 \\
& Vid2WAM & \textbf{2.0} & \textbf{2.0} & \textbf{4.0} & \textbf{1.0} & \textbf{2.3} \\
\bottomrule
\end{tabular}%
}
\normalsize
\addtocounter{table}{2}
\caption{Success rates (\%) evaluated only on the withheld LIBERO target tasks.
Each split removes the listed two task IDs from every suite.}
\label{tab:libero_novel_success}
\addtocounter{table}{-3}
\end{table}

Results show that Vid2WAM improves average withheld-target success for all three splits on both
benchmarks. The "zero to one" improvements on most suites indicate that our distillation framework can adjust the task execution process correctly. Low absolute success on several Goal and Long targets nevertheless
shows that adaptation without target-task action trajectories remains somewhat difficult.
\par

\FloatBarrier

\begin{table*}[!t]
\centering
\scriptsize
\setlength{\tabcolsep}{2.8pt}
\resizebox{\textwidth}{!}{%
\begin{tabular}{ll|cccc|ccccccc|c}
\toprule
\multirow{2}{*}{Novel IDs} & \multirow{2}{*}{Model}
& \multicolumn{4}{c|}{Suite}
& \multicolumn{7}{c|}{Perturbation}
& \multirow{2}{*}{Overall} \\
\cmidrule(lr){3-6} \cmidrule(lr){7-13}
& & Spatial & Object & Goal & Long
& Cam. & Robot & Lang. & Light & BG & Noise & Layout & \\
\midrule
\multirow{2}{*}{$0/1$}
& Fast-WAM & 58.4 & 61.2 & \textbf{39.4} & 37.4
& \textbf{28.2} & 36.1 & 60.0 & 70.5 & \textbf{54.5} & 39.6 & \textbf{58.7} & 49.1 \\
& Vid2WAM & \textbf{60.2} & \textbf{63.2} & 37.2 & \textbf{39.8}
& 24.8 & \textbf{40.0} & \textbf{66.3} & \textbf{76.5} & 48.4 & \textbf{40.6} & \textbf{58.7} & \textbf{50.1} \\
\midrule
\multirow{2}{*}{$3/6$}
& Fast-WAM & 40.6 & \textbf{56.6} & 30.2 & \textbf{36.4}
& 13.2 & 37.5 & 52.3 & 64.2 & \textbf{43.0} & \textbf{32.9} & 48.4 & 41.0 \\
& Vid2WAM & \textbf{45.0} & 53.6 & \textbf{40.2} & 35.0
& \textbf{15.1} & \textbf{38.3} & \textbf{63.2} & \textbf{70.9} & 40.1 & 31.8 & \textbf{49.8} & \textbf{43.5} \\
\midrule
\multirow{2}{*}{$8/9$}
& Fast-WAM & 48.0 & 59.6 & 37.0 & \textbf{38.6}
& 24.1 & 42.5 & 64.2 & 67.2 & 39.0 & 34.6 & \textbf{52.7} & 45.8 \\
& Vid2WAM & \textbf{52.4} & \textbf{60.2} & \textbf{41.4} & \textbf{38.6}
& \textbf{28.5} & \textbf{42.8} & \textbf{67.7} & \textbf{67.5} & \textbf{39.7} & \textbf{42.4} & 51.6 & \textbf{48.1} \\
\bottomrule
\end{tabular}%
}
\normalsize
\caption{Sensitivity to novel-task selection on LIBERO-Plus. Each split
treats the listed two task IDs in every suite as novel; $8/9$ is the main
novel setting reported in the main paper. Best results within each split are
bolded.}
\label{tab:libero_plus_split_sensitivity}
\end{table*}

\begin{table*}[!t]
\begin{minipage}[t]{0.485\textwidth}
\centering
\small
\setlength{\tabcolsep}{3pt}
\resizebox{\linewidth}{!}{%
\begin{tabular}{llccccc}
\toprule
Novel IDs & Method & Spatial & Object & Goal & Long & \textbf{Avg.} \\
\midrule
\multirow{2}{*}{$0/1$}
& Fast-WAM & 82.4 & 79.8 & \textbf{76.4} & \textbf{75.6} & 78.6 \\
& Vid2WAM & \textbf{94.2} & \textbf{92.6} & 76.2 & 71.8 & \textbf{83.7} \\
\midrule
\multirow{2}{*}{$3/6$}
& Fast-WAM & 73.0 & 79.2 & \textbf{77.6} & \textbf{76.4} & 76.6 \\
& Vid2WAM & \textbf{75.6} & \textbf{84.4} & 76.2 & 74.2 & \textbf{77.6} \\
\midrule
\multirow{2}{*}{$8/9$}
& Fast-WAM & 76.8 & 79.0 & 77.8 & 73.0 & 76.7 \\
& Vid2WAM & \textbf{79.4} & \textbf{79.4} & \textbf{79.2}
& \textbf{75.2} & \textbf{78.3} \\
\bottomrule
\end{tabular}%
}
\normalsize
\caption{Sensitivity to novel-task selection on LIBERO. Best results within each split are bolded.}
\label{tab:libero_different_task}
\end{minipage}\hfill
\begin{minipage}[t]{0.485\textwidth}
\centering
\small
\setlength{\tabcolsep}{2.5pt}
\resizebox{\linewidth}{!}{%
\begin{tabular}{llccccc}
\toprule
Novel IDs & Method & Spatial & Object & Goal & Long & \textbf{Avg.} \\
\midrule
\multirow{2}{*}{$0/1$}
& Fast-WAM & 51.0 & 1.0 & 0.0 & 0.0 & 13.0 \\
& Vid2WAM & \textbf{69.0} & \textbf{34.0} & 0.0 & 0.0 & \textbf{25.8} \\
\midrule
\multirow{2}{*}{$3/6$}
& Fast-WAM & 0.0 & 0.0 & 1.0 & 0.0 & 0.3 \\
& Vid2WAM & \textbf{2.0} & \textbf{9.0} & \textbf{2.0} & 0.0 & \textbf{3.3} \\
\midrule
\multirow{2}{*}{$8/9$}
& Fast-WAM & 1.0 & 2.0 & 1.0 & 0.0 & 1.0 \\
& Vid2WAM & \textbf{2.0} & \textbf{12.0} & \textbf{5.0} & 0.0 & \textbf{4.8} \\
\bottomrule
\end{tabular}%
}
\normalsize
\addtocounter{table}{1}
\caption{Success rates (\%) evaluated only on the withheld LIBERO-Plus target tasks.
Each split removes the listed two task IDs from every suite.}
\label{tab:libero_plus_novel_success}
\end{minipage}
\end{table*}

\newpage

%% file: supp/sections/realworld_evaluation.tex
\raggedbottom

\section{Real-World Experiment Details}

Figure~\ref{fig:realworld_setup} illustrates the real-world platform used for
data collection and evaluation. A camera-equipped dual-arm robot is
operated through the paired teleoperation interface. The top-down camera,
studio lighting, and reconfigurable tabletop provide controlled variation in
viewpoint, illumination, texture, object layout, and task composition.

\subsection{Detailed Task Description}

The evaluation includes six seen tasks with expert trajectories and
three novel tasks trained only from true or synthetic initial observations, language instructions, and offline
teacher supervision. The novel tasks introduce behaviors or object--action
compositions absent from the labeled trajectories.

\paragraph{Seen tasks.}
\begin{itemize}
    \item \textbf{Click Bell:} the right gripper closes and presses a tabletop
    bell.
    \item \textbf{Item Handover:} the right arm lifts a pen upright, transfers
    it to the left arm, and the left arm drops it into a pen holder.
    \item \textbf{Carry Basket:} the left arm places a milk carton into a
    basket, after which the right arm lifts and relocates the basket by its
    handle.
    \item \textbf{Open Pan Place:} the left arm opens the pan lid, the right arm
    places a carrot inside, and the left arm closes the lid.
    \item \textbf{Open Drawer Place:} the right arm opens a handled drawer while
    the left arm places a tomato inside.
    \item \textbf{Insert Test Tube:} the arms hand over a transparent test tube
    before vertically inserting it into a rack.
\end{itemize}

\paragraph{Novel tasks.}
\begin{itemize}
    \item \textbf{Pick Test Tube:} the right arm grasps a test tube already
    inserted in the rack and pulls it out.
    \item \textbf{Take Tissue:} the right arm pinches and extracts a tissue from
    a soft tissue pack.
    \item \textbf{Tomato Basket:} the left arm picks up a tomato and place it into the basket.
\end{itemize}

\subsection{Initial-State Augmentation}

\begin{figure*}[p]
\centering
\includegraphics[width=0.85\textwidth,trim=0 0 0 3pt,clip]{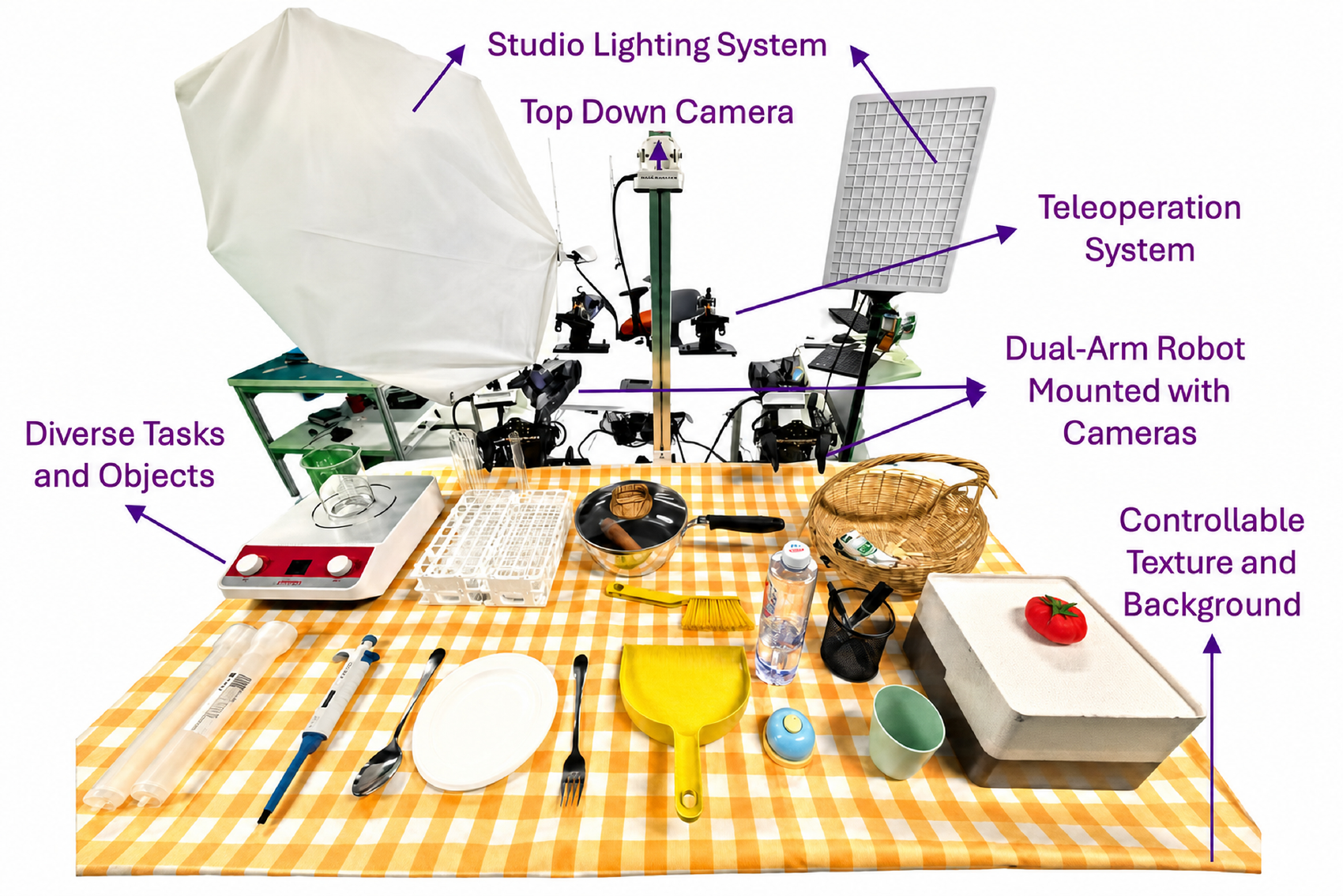}
\caption{Real-world bimanual platform, including the teleoperation interface,
camera-equipped dual-arm robot, top-down camera, studio lighting, and
reconfigurable tabletop with task objects.}
\label{fig:realworld_setup}
\par\medskip
\includegraphics[width=0.85\textwidth]{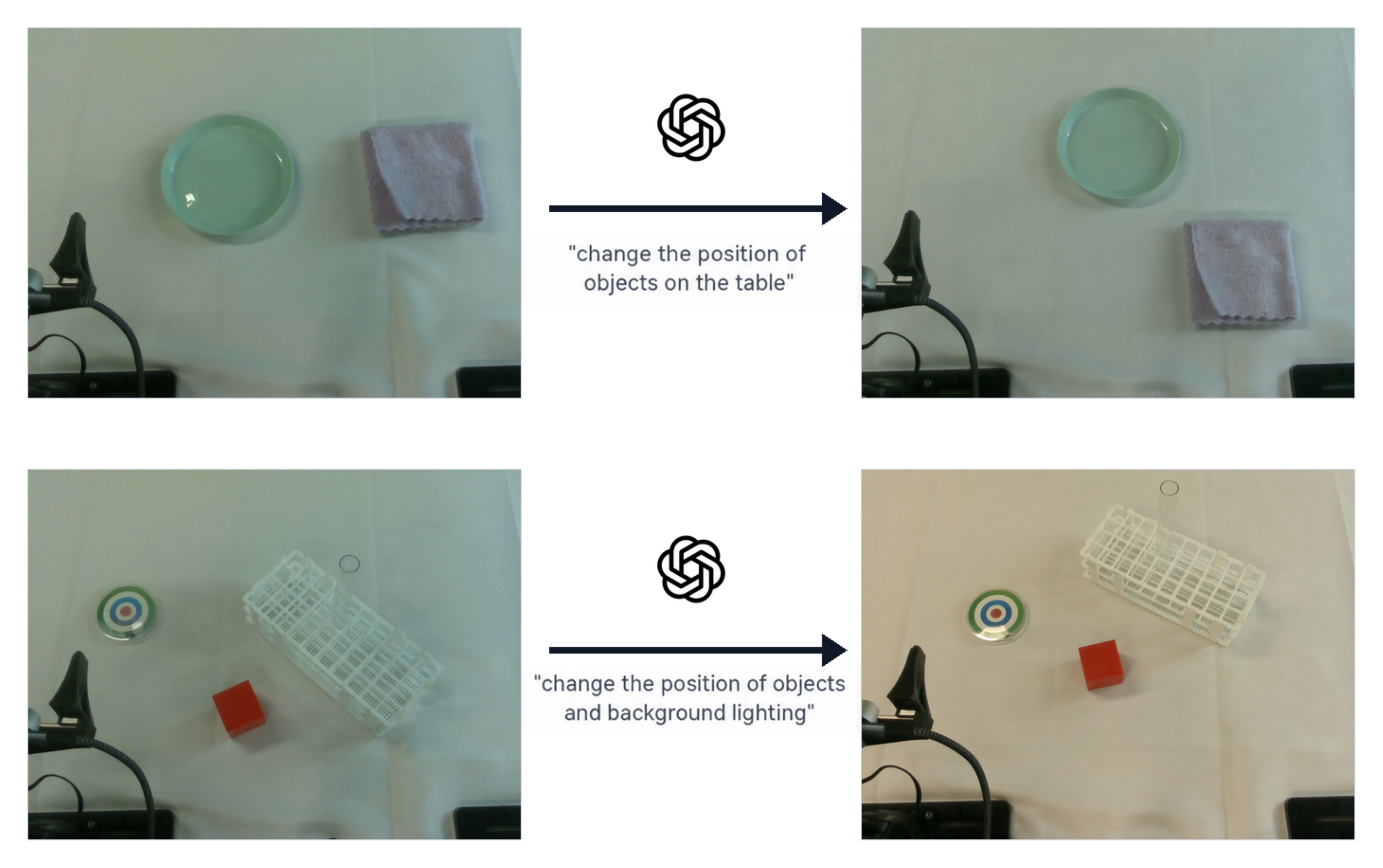}
\caption{Real-world initial-state augmentation with GPT-Image-2. In each row,
the collected overhead observation (left) is edited into a new conditioning
image (right). The top example changes object placement, while the bottom
example changes both object placement and illumination.}
\label{fig:realworld_imagegen}
\end{figure*}

As shown in Figure~\ref{fig:realworld_imagegen}, for each novel task, GPT-Image-2 generates 30 synthetic images from a
collected overhead observation in clean setting with only task-specific objects, and 30 images from an observation with different distractors. The generated images
vary object placement and illumination without requiring additional real frames.

\subsection{Spatial and Temporal Masking}
\label{app:loss_masking}
Synthesis is restricted to the overhead view because independently generated
views do not reliably preserve cross-view alignment. To fit the model, the
generated image occupies the upper camera slot, while unavailable wrist-camera regions are zero-filled, and a spatial latent
mask limits pseudo-video loss to the valid region. Ground-truth multi-view and
action supervision remain unmasked.
\paragraph{Spatial masking.}
% The spatial mask is used for pseudo videos whose available camera view occupies
% only part of the multi-view canvas.  
In the single-view construction, the
  overhead image is resized to the top $224\times448$ region of a
$448\times448$ composite image, while the bottom $224\times448$ region is a
black placeholder.  The student VAE has a spatial downsampling factor of 16,
so this image is represented on a $28\times28$ latent grid.  Consequently,
latent rows $h\in[0,14)$ and all columns $w\in[0,28)$ are valid, whereas rows
$h\in[14,28)$ are ignored.  Multi-view samples that do not contain
an explicit spatial mask are assigned an all-one mask and therefore retain
the original full-image loss.

\paragraph{Temporal masking.}
Video clips are padded to a fixed training length before VAE encoding. Because
the student VAE downsamples time by a factor of four, frame-level padding flags
are converted into a latent-time validity mask. A latent step is ignored only
when all contributing video frames are padding, and the video loss is averaged
over the remaining valid latent steps. The first frame, which provides the image
condition, is handled separately. This temporal masking is the same for
single-view and multi-view samples; single-view pseudo videos additionally use
the spatial mask described above. Action chunks are fitted to the student
horizon, with padded action positions excluded from supervision.

\paragraph{Rollout-duration cutoff.}
Simulation pseudo-rollouts are truncated to the first 8 seconds of video before
constructing pseudo-data to mitigate long-horizon generation errors. Real-world
pseudo-rollouts use all frames from the 20-second generated video and do not
apply this temporal cutoff, which improves real-world task success rates. This
duration cutoff is separate from the padding-based temporal mask above.

\flushbottom

%% file: supp/sections/robotwin_analysis.tex
\section{Full RoboTwin Task-Level Results}

Tables~\ref{tab:robotwin_low1010_comparison} and
\ref{tab:robotwin_novel_comparison} expand the main paper's aggregate RoboTwin
results to all 50 tasks under clean and randomized evaluation settings. Vid2WAM achieves the best performance on low-data and novel regimes under both evaluation conditions.
Table~\ref{tab:robotwin_novel_comparison} also marks the 15 withheld target tasks in
bold. Vid2WAM leads both full-benchmark and novel-subset aggregates under clean
and randomized settings, with larger gains on the novel subset.

\begin{table*}[!t]
\centering
\tiny
\setlength{\tabcolsep}{2.8pt}
\renewcommand{\arraystretch}{0.65}
\resizebox{0.98\textwidth}{!}{%
\begin{tabular}{lcccccccc}
\toprule
\multirow{2}{*}{Task}
& \multicolumn{2}{c}{Vid2WAM}
& \multicolumn{2}{c}{Fast-WAM}
& \multicolumn{2}{c}{$\pi_{0.5}$}
& \multicolumn{2}{c}{Motus} \\
\cmidrule(lr){2-3}
\cmidrule(lr){4-5}
\cmidrule(lr){6-7}
\cmidrule(lr){8-9}
& Clean & Random
& Clean & Random
& Clean & Random
& Clean & Random \\
\midrule
adjust\_bottle
& 0.95 & \textbf{0.90} & \textbf{0.99} & 0.86 & 0.94 & 0.88 & 0.85 & 0.87 \\

beat\_block\_hammer
& \textbf{0.84} & \textbf{0.66} & 0.69 & 0.52 & 0.75 & 0.64 & 0.48 & 0.29 \\

blocks\_ranking\_rgb
& \textbf{0.42} & \textbf{0.16} & 0.39 & 0.13 & 0.13 & 0.15 & 0.27 & \textbf{0.16} \\

blocks\_ranking\_size
& \textbf{0.34} & 0.11 & 0.16 & 0.08 & 0.10 & 0.09 & 0.19 & \textbf{0.17} \\

click\_alarmclock
& 0.82 & 0.69 & 0.78 & 0.51 & 0.79 & 0.84 & \textbf{0.93} & \textbf{0.94} \\

click\_bell
& 0.96 & 0.84 & 0.95 & 0.85 & 0.92 & \textbf{0.94} & \textbf{0.97} & 0.88 \\

dump\_bin\_bigbin
& \textbf{0.90} & \textbf{0.77} & 0.88 & 0.74 & 0.69 & 0.47 & 0.62 & 0.51 \\

grab\_roller
& \textbf{0.99} & \textbf{0.91} & \textbf{0.99} & 0.89 & 0.92 & 0.74 & 0.98 & 0.73 \\

handover\_block
& \textbf{0.30} & \textbf{0.20} & 0.13 & 0.05 & 0.05 & 0.04 & 0.09 & 0.02 \\

handover\_mic
& \textbf{0.96} & \textbf{0.63} & 0.95 & 0.62 & 0.32 & 0.20 & 0.46 & 0.44 \\

lift\_pot
& 0.14 & 0.17 & 0.18 & 0.20 & \textbf{0.59} & \textbf{0.39} & 0.14 & 0.09 \\

move\_can\_pot
& 0.49 & \textbf{0.45} & 0.57 & 0.33 & \textbf{0.60} & 0.41 & 0.08 & 0.25 \\

move\_playingcard\_away
& 0.78 & 0.68 & 0.73 & 0.59 & 0.81 & 0.86 & \textbf{0.92} & \textbf{0.87} \\

move\_stapler\_pad
& 0.03 & 0.02 & 0.01 & 0.00 & \textbf{0.18} & \textbf{0.11} & 0.08 & 0.04 \\

hanging\_mug
& 0.07 & \textbf{0.11} & 0.09 & 0.10 & 0.02 & 0.03 & \textbf{0.17} & 0.06 \\

open\_laptop
& 0.82 & 0.69 & \textbf{0.84} & 0.63 & 0.71 & \textbf{0.72} & 0.70 & \textbf{0.72} \\

open\_microwave
& 0.30 & 0.18 & 0.24 & 0.18 & 0.05 & 0.09 & \textbf{0.57} & \textbf{0.40} \\

pick\_diverse\_bottles
& 0.25 & 0.11 & 0.20 & 0.03 & 0.44 & 0.23 & \textbf{0.45} & \textbf{0.33} \\

pick\_dual\_bottles
& 0.29 & 0.15 & 0.25 & 0.12 & \textbf{0.59} & \textbf{0.51} & 0.58 & 0.38 \\

place\_a2b\_left
& 0.30 & 0.26 & 0.24 & 0.09 & 0.49 & 0.42 & \textbf{0.77} & \textbf{0.60} \\

place\_a2b\_right
& 0.34 & 0.33 & 0.21 & 0.14 & 0.45 & 0.37 & \textbf{0.77} & \textbf{0.57} \\

place\_bread\_basket
& \textbf{0.70} & 0.56 & 0.60 & 0.51 & 0.61 & \textbf{0.58} & 0.54 & 0.54 \\

place\_bread\_skillet
& 0.45 & 0.29 & 0.55 & 0.23 & 0.36 & 0.24 & \textbf{0.57} & \textbf{0.44} \\

place\_can\_basket
& 0.68 & 0.52 & \textbf{0.79} & 0.51 & 0.62 & \textbf{0.54} & 0.29 & 0.32 \\

place\_cans\_plasticbox
& \textbf{0.88} & \textbf{0.69} & \textbf{0.88} & \textbf{0.69} & 0.52 & 0.42 & 0.28 & 0.18 \\

place\_container\_plate
& 0.96 & \textbf{0.92} & \textbf{0.98} & 0.87 & 0.95 & 0.82 & 0.91 & 0.79 \\

place\_dual\_shoes
& 0.15 & 0.11 & \textbf{0.22} & \textbf{0.13} & 0.12 & 0.06 & 0.19 & 0.02 \\

place\_empty\_cup
& 0.82 & 0.64 & 0.69 & 0.48 & 0.77 & 0.80 & \textbf{0.88} & \textbf{0.86} \\

place\_fan
& 0.18 & 0.13 & \textbf{0.30} & 0.13 & 0.08 & 0.14 & 0.18 & \textbf{0.15} \\

place\_burger\_fries
& \textbf{0.84} & 0.77 & 0.79 & 0.69 & 0.81 & \textbf{0.82} & 0.57 & 0.58 \\

place\_mouse\_pad
& 0.18 & 0.07 & 0.09 & 0.03 & \textbf{0.28} & 0.14 & 0.18 & \textbf{0.29} \\

place\_object\_basket
& 0.70 & 0.59 & \textbf{0.72} & \textbf{0.69} & 0.31 & 0.17 & 0.59 & 0.61 \\

place\_object\_scale
& 0.36 & 0.28 & 0.24 & 0.13 & 0.45 & \textbf{0.42} & \textbf{0.49} & 0.35 \\

place\_object\_stand
& \textbf{0.88} & \textbf{0.80} & 0.81 & 0.52 & 0.75 & 0.61 & 0.60 & 0.56 \\

place\_phone\_stand
& \textbf{0.60} & \textbf{0.60} & 0.51 & 0.36 & 0.47 & 0.47 & 0.38 & 0.20 \\

move\_pillbottle\_pad
& 0.37 & 0.22 & 0.31 & 0.20 & \textbf{0.54} & \textbf{0.58} & 0.36 & 0.15 \\

place\_shoe
& 0.37 & 0.21 & 0.37 & 0.34 & 0.24 & 0.21 & \textbf{0.57} & \textbf{0.61} \\

press\_stapler
& 0.92 & 0.81 & 0.87 & 0.68 & 0.80 & 0.71 & \textbf{1.00} & \textbf{0.91} \\

put\_bottles\_dustbin
& \textbf{0.13} & \textbf{0.15} & 0.08 & 0.04 & 0.03 & 0.01 & 0.07 & 0.00 \\

put\_object\_cabinet
& \textbf{0.34} & \textbf{0.22} & 0.29 & 0.20 & 0.09 & 0.06 & 0.10 & 0.19 \\

rotate\_qrcode
& 0.53 & \textbf{0.33} & \textbf{0.54} & 0.22 & 0.32 & 0.21 & 0.24 & 0.22 \\

scan\_object
& 0.31 & 0.21 & 0.26 & 0.24 & 0.28 & 0.25 & \textbf{0.59} & \textbf{0.49} \\

shake\_bottle
& \textbf{1.00} & \textbf{1.00} & \textbf{1.00} & 0.98 & 0.96 & 0.97 & 0.94 & 0.94 \\

shake\_bottle\_horizontally
& \textbf{1.00} & \textbf{1.00} & \textbf{1.00} & 0.97 & 0.97 & 0.97 & 0.93 & 0.91 \\

stack\_blocks\_three
& \textbf{0.30} & \textbf{0.18} & 0.18 & 0.10 & 0.05 & 0.07 & 0.00 & 0.03 \\

stack\_blocks\_two
& 0.66 & \textbf{0.54} & 0.68 & 0.43 & 0.60 & 0.42 & \textbf{0.70} & \textbf{0.54} \\

stack\_bowls\_three
& \textbf{0.62} & \textbf{0.54} & 0.55 & 0.39 & 0.27 & 0.13 & 0.31 & 0.22 \\

stack\_bowls\_two
& \textbf{0.89} & \textbf{0.85} & 0.81 & 0.69 & 0.73 & 0.71 & 0.79 & 0.80 \\

stamp\_seal
& 0.35 & 0.25 & 0.25 & 0.16 & \textbf{0.41} & \textbf{0.43} & 0.10 & 0.09 \\

turn\_switch
& 0.22 & 0.18 & 0.24 & 0.26 & 0.40 & 0.31 & \textbf{0.54} & \textbf{0.49} \\

\midrule
\textbf{Overall}
& \textbf{0.5536} & \textbf{0.4536}
& 0.5214 & 0.3906
& 0.4866 & 0.4280
& 0.4992 & 0.4360 \\
\bottomrule
\end{tabular}%
}
\normalsize
\caption{Success rates on the RoboTwin benchmark (low-data regime).}
\label{tab:robotwin_low1010_comparison}
\end{table*}

\begin{table*}[!t]
\centering
\tiny
\setlength{\tabcolsep}{2.8pt}
\renewcommand{\arraystretch}{0.65}
\resizebox{0.98\textwidth}{!}{%
\begin{tabular}{lcccccccc}
\toprule
\multirow{2}{*}{Task}
& \multicolumn{2}{c}{Vid2WAM}
& \multicolumn{2}{c}{Fast-WAM}
& \multicolumn{2}{c}{$\pi_{0.5}$}
& \multicolumn{2}{c}{Motus} \\
\cmidrule(lr){2-3}
\cmidrule(lr){4-5}
\cmidrule(lr){6-7}
\cmidrule(lr){8-9}
& Clean & Random
& Clean & Random
& Clean & Random
& Clean & Random \\
\midrule
adjust\_bottle
& \textbf{1.00} & \textbf{1.00}
& \textbf{1.00} & 0.99
& 0.99 & 0.98
& \textbf{1.00} & \textbf{1.00} \\

beat\_block\_hammer
& 0.99 & 0.96
& \textbf{1.00} & \textbf{0.99}
& 0.89 & 0.89
& 0.83 & 0.85 \\

\textbf{blocks\_ranking\_rgb}
& \textbf{0.08} & 0.02
& 0.06 & \textbf{0.08}
& 0.03 & 0.04
& 0.04 & \textbf{0.08} \\

blocks\_ranking\_size
& 0.83 & 0.84
& \textbf{0.85} & 0.82
& 0.69 & 0.42
& 0.79 & \textbf{0.87} \\

\textbf{click\_alarmclock}
& 0.55 & \textbf{0.81}
& 0.50 & 0.44
& \textbf{0.58} & 0.53
& 0.49 & 0.79 \\

click\_bell
& \textbf{1.00} & \textbf{1.00}
& \textbf{1.00} & \textbf{1.00}
& 0.91 & 0.79
& \textbf{1.00} & \textbf{1.00} \\

dump\_bin\_bigbin
& \textbf{0.98} & \textbf{0.94}
& \textbf{0.98} & 0.93
& 0.88 & 0.89
& 0.94 & \textbf{0.94} \\

grab\_roller
& \textbf{1.00} & \textbf{1.00}
& \textbf{1.00} & \textbf{1.00}
& \textbf{1.00} & \textbf{1.00}
& 0.99 & 0.99 \\

handover\_block
& \textbf{0.95} & 0.86
& \textbf{0.95} & 0.85
& 0.78 & 0.72
& 0.92 & \textbf{0.89} \\

handover\_mic
& \textbf{1.00} & \textbf{1.00}
& 0.99 & 0.99
& 0.97 & 0.83
& \textbf{1.00} & \textbf{1.00} \\

lift\_pot
& 0.98 & 0.99
& \textbf{0.99} & \textbf{1.00}
& 0.80 & 0.86
& 0.97 & 0.96 \\

move\_can\_pot
& 0.94 & 0.98
& \textbf{0.98} & \textbf{0.99}
& 0.48 & 0.56
& 0.93 & 0.97 \\

\textbf{move\_playingcard\_away}
& 0.04 & 0.02
& \textbf{0.11} & 0.00
& 0.08 & \textbf{0.08}
& 0.06 & 0.01 \\

move\_stapler\_pad
& 0.72 & \textbf{0.73}
& 0.67 & 0.65
& 0.63 & 0.55
& \textbf{0.77} & 0.71 \\

hanging\_mug
& \textbf{0.66} & \textbf{0.51}
& 0.49 & 0.47
& 0.08 & 0.12
& 0.51 & 0.44 \\

\textbf{open\_laptop}
& \textbf{0.18} & \textbf{0.14}
& 0.04 & 0.03
& 0.12 & \textbf{0.14}
& 0.11 & 0.10 \\

open\_microwave
& 0.47 & 0.48
& 0.55 & 0.55
& 0.53 & 0.65
& \textbf{0.77} & \textbf{0.72} \\

pick\_diverse\_bottles
& 0.68 & 0.76
& \textbf{0.78} & \textbf{0.86}
& 0.74 & 0.66
& 0.48 & 0.61 \\

\textbf{pick\_dual\_bottles}
& 0.72 & 0.76
& \textbf{0.99} & \textbf{0.91}
& 0.79 & 0.81
& 0.65 & 0.72 \\

place\_a2b\_left
& \textbf{0.96} & 0.93
& 0.94 & \textbf{0.94}
& 0.90 & 0.87
& 0.94 & 0.92 \\

\textbf{place\_a2b\_right}
& 0.10 & \textbf{0.17}
& 0.03 & 0.10
& \textbf{0.16} & 0.15
& 0.15 & 0.15 \\

\textbf{place\_bread\_basket}
& \textbf{0.71} & \textbf{0.66}
& 0.01 & 0.00
& 0.19 & 0.21
& 0.29 & 0.23 \\

place\_bread\_skillet
& \textbf{0.96} & 0.85
& 0.95 & \textbf{0.95}
& 0.78 & 0.79
& 0.89 & 0.87 \\

place\_can\_basket
& 0.67 & 0.70
& 0.71 & 0.58
& \textbf{0.74} & \textbf{0.73}
& 0.64 & 0.72 \\

place\_cans\_plasticbox
& \textbf{0.99} & \textbf{0.98}
& 0.98 & 0.96
& 0.82 & 0.75
& 0.98 & 0.97 \\

\textbf{place\_container\_plate}
& \textbf{0.97} & 0.90
& 0.73 & 0.61
& 0.81 & 0.86
& 0.95 & \textbf{0.92} \\

place\_dual\_shoes
& 0.88 & \textbf{0.93}
& \textbf{0.92} & 0.91
& 0.68 & 0.71
& 0.81 & 0.85 \\

place\_empty\_cup
& \textbf{1.00} & \textbf{1.00}
& 0.99 & \textbf{1.00}
& \textbf{1.00} & \textbf{1.00}
& \textbf{1.00} & \textbf{1.00} \\

place\_fan
& \textbf{0.97} & \textbf{0.96}
& \textbf{0.97} & 0.95
& 0.83 & 0.85
& \textbf{0.97} & 0.94 \\

\textbf{place\_burger\_fries}
& \textbf{0.17} & 0.17
& 0.02 & 0.00
& 0.03 & 0.02
& 0.16 & \textbf{0.21} \\

place\_mouse\_pad
& 0.90 & 0.86
& \textbf{0.94} & \textbf{0.93}
& 0.67 & 0.74
& 0.89 & 0.92 \\

place\_object\_basket
& 0.74 & 0.86
& 0.81 & 0.81
& 0.78 & 0.82
& \textbf{0.90} & \textbf{0.91} \\

place\_object\_scale
& 0.90 & 0.94
& 0.89 & \textbf{0.96}
& 0.77 & 0.75
& \textbf{0.92} & 0.93 \\

\textbf{place\_object\_stand}
& \textbf{0.62} & \textbf{0.63}
& 0.29 & 0.34
& 0.24 & 0.22
& 0.41 & 0.52 \\

place\_phone\_stand
& \textbf{0.98} & \textbf{0.99}
& \textbf{0.98} & \textbf{0.99}
& 0.83 & 0.87
& 0.89 & 0.90 \\

\textbf{move\_pillbottle\_pad}
& 0.11 & 0.05
& 0.01 & 0.01
& \textbf{0.18} & \textbf{0.15}
& 0.05 & 0.04 \\

place\_shoe
& 0.94 & \textbf{1.00}
& 0.97 & 0.99
& 0.93 & 0.94
& \textbf{0.98} & 0.99 \\

\textbf{press\_stapler}
& 0.99 & \textbf{1.00}
& \textbf{1.00} & 0.97
& 0.94 & 0.93
& \textbf{1.00} & \textbf{1.00} \\

put\_bottles\_dustbin
& \textbf{0.83} & \textbf{0.93}
& 0.80 & 0.89
& 0.49 & 0.42
& 0.56 & 0.72 \\

put\_object\_cabinet
& 0.93 & \textbf{0.91}
& \textbf{0.94} & \textbf{0.91}
& 0.76 & 0.77
& 0.88 & 0.86 \\

rotate\_qrcode
& 0.93 & 0.85
& \textbf{0.96} & \textbf{0.90}
& 0.81 & 0.84
& 0.89 & 0.85 \\

scan\_object
& \textbf{0.94} & 0.89
& 0.91 & \textbf{0.95}
& 0.73 & 0.64
& 0.82 & 0.91 \\

shake\_bottle
& \textbf{1.00} & \textbf{1.00}
& \textbf{1.00} & 0.99
& 0.99 & 0.97
& \textbf{1.00} & \textbf{1.00} \\

\textbf{shake\_bottle\_horizontally}
& \textbf{1.00} & 0.99
& \textbf{1.00} & 0.99
& \textbf{1.00} & \textbf{1.00}
& \textbf{1.00} & \textbf{1.00} \\

stack\_blocks\_three
& \textbf{0.97} & 0.95
& 0.96 & \textbf{0.96}
& 0.86 & 0.84
& 0.94 & 0.91 \\

\textbf{stack\_blocks\_two}
& \textbf{1.00} & \textbf{1.00}
& \textbf{1.00} & 0.97
& 0.97 & 0.98
& \textbf{1.00} & \textbf{1.00} \\

stack\_bowls\_three
& 0.81 & \textbf{0.82}
& 0.77 & 0.80
& \textbf{0.84} & 0.66
& 0.83 & 0.74 \\

\textbf{stack\_bowls\_two}
& 0.96 & \textbf{0.97}
& 0.96 & \textbf{0.97}
& 0.86 & 0.82
& \textbf{0.98} & 0.96 \\

stamp\_seal
& 0.77 & \textbf{0.86}
& \textbf{0.84} & 0.77
& 0.71 & 0.73
& 0.71 & 0.65 \\

turn\_switch
& 0.68 & 0.69
& 0.65 & 0.72
& 0.69 & 0.69
& \textbf{0.86} & \textbf{0.90} \\

\midrule
\textbf{Overall}
& \textbf{0.7830} & \textbf{0.7848}
& 0.7572 & 0.7474
& 0.6792 & 0.6648
& 0.7508 & 0.7628 \\

\textbf{Novel subset}
& \textbf{0.5467} & \textbf{0.5527}
& 0.4500 & 0.4280
& 0.4653 & 0.4627
& 0.4893 & 0.5153 \\
\bottomrule
\end{tabular}%
}
\normalsize
\caption{Success rates on the RoboTwin benchmark (novel regime). Novel target tasks are marked in bold.}
\label{tab:robotwin_novel_comparison}
\end{table*}